%% file: paper_template.tex
\documentclass[conference]{IEEEtran}
\usepackage{times}
\usepackage[numbers]{natbib}
\usepackage{multicol}
\usepackage[bookmarks=true]{hyperref}
\usepackage{graphics} 
\usepackage{mathptmx}
\usepackage{color}
\usepackage[dvipsnames]{xcolor}
\usepackage{graphicx}
\usepackage[ruled]{algorithm2e}
\usepackage{epsfig}
\usepackage{amsmath}
\usepackage{amssymb}
\usepackage{wrapfig,lipsum,booktabs}
\usepackage{subfig}
\usepackage{multirow}
\usepackage{utfsym}

\usepackage[hmargin=2cm,vmargin=2.5cm]{geometry}
\usepackage{etoolbox}
\usepackage{booktabs}
\AtBeginEnvironment{longtable}{%
  \addfontfeature{RawFeature=+tnum;-onum}
}

\usepackage{colortbl}%
\newcommand{\myrowcolour}{\rowcolor[gray]{0.925}}
\newcommand{\highest}[1]{\textcolor{Maroon}{\textbf{#1}}}%
\newcommand{\yes}{\large \color{OliveGreen}\checkmark}
\newcommand{\no}{\color{BrickRed} \scalebox{1}{\usym{2613}}}
\makeatletter
\def\blfootnote{\xdef\@thefnmark{}\@footnotetext}
\makeatother

\newcommand{\ours}{AnyTeleop }
\newcommand{\oursnospace}{AnyTeleop}
\newcommand{\etal}{\textit{et al.}}

\newcommand{\CC}{\cellcolor[gray]{0.925}}

\newlength\savewidth

\begin{document}

\title{AnyTeleop: A General Vision-Based Dexterous Robot Arm-Hand Teleoperation System}

\author{Yuzhe Qin$^{1}$, \quad
Wei Yang$^{2}$, \quad
Binghao Huang$^{1}$, \quad
Karl Van Wyk$^{2}$ \quad
\\ Hao Su$^{1}$,  \quad
Xiaolong Wang$^{1}$, \quad
Yu-Wei Chao$^{2}$, \quad
Dieter Fox$^{2}$ \\
$^{1}$UC San Diego \hspace{0.2in}
$^{2}$NVIDIA \hspace{0.2in} \\
{\color{blue}{\texttt{\url{https://yzqin.github.io/anyteleop/}}}}
}

\twocolumn[{%
\renewcommand\twocolumn[1][]{#1}%
\maketitle
\begin{center}
    \centering 
    \includegraphics[width=\linewidth]{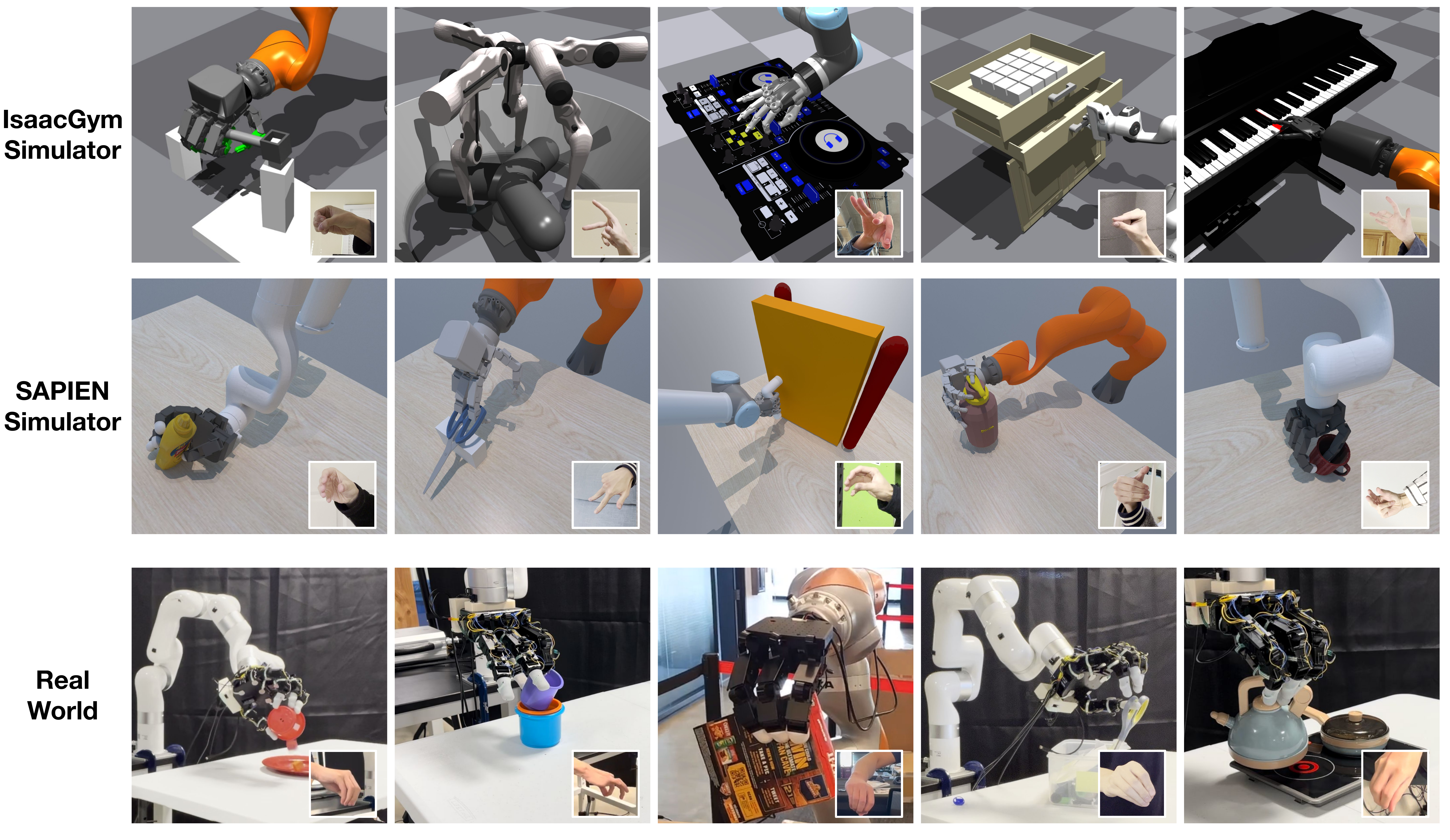}
    \captionof{figure}{We present \oursnospace, a vision-based teleoperation system for a variety of scenarios to solve a wide range of manipulation tasks. \ours can be used for various robot arms with different robot hands. It also supports teleoperation within different realities, such as IsaacGym (top row), and SAPIEN simulator (middle row), and real world (bottom rows). }
    \label{fig:teaser}
\end{center}
}]
{\blfootnote{{Yuzhe Qin was an intern at NVIDIA during the project.}}}
\begin{abstract}
Vision-based teleoperation offers the possibility to endow robots with human-level intelligence to physically interact with the environment, while only requiring low-cost camera sensors. However, current vision-based teleoperation systems are designed and engineered towards a particular robot model and deploy environment, which scales poorly as the pool of the robot models expands and the variety of the operating environment increases. In this paper, we propose \oursnospace, a unified and general teleoperation system to support multiple different arms, hands, realities, and camera configurations within a single system. Although being designed to provide great flexibility to the choice of simulators and real hardware, our system can still achieve great performance. For real-world experiments, \ours
can outperform a previous system that was designed for a specific robot hardware with a higher success rate, using the same robot. For teleoperation in simulation, \ours leads to better imitation learning performance, compared with a previous system that is particularly designed for that simulator.
\end{abstract}

\IEEEpeerreviewmaketitle

\input{sections/intro.tex}
\input{sections/related.tex}

\input{sections/overview.tex}
\input{sections/teleop_system.tex}
\input{sections/web_visualization.tex}
\input{sections/system_eval.tex}

\input{sections/application.tex}

\section{Failure Modes}
As illustrated on the project page, we have identified two failure modes: (i) loss of tracking during fast human hand motion, which triggers a pause and re-detection process; (ii) unreliable hand pose when the hand is in self-occlusion. For mode (i), the workaround is to instruct the operator to slow down their hand motion. The issue (ii) can be solved by incorporating multiple cameras for tasks that require significant hand rotation.

\section{Conclusion}
In this paper, we introduced \oursnospace, a versatile teleoperation system that can be applied to diverse robots, assorted reality, and varied camera setup, and can be operated by a flexible number of users from any geographic locations. The experiments show that \ours outperforms previous systems in both simulation and real-world scenarios while offering superior generalizability and flexibility. Our commitment to an open-source approach will facilitate further research in the field of teleoperation.

\section{Acknowledgement}
We express our gratitude to Ankur Handa, Balakumar Sundaralingam, and Nick Walker for their insightful discussions throughout the development process of the motion control module in AnyTeleop. We would also like to extend our thanks to Isabella Liu, An-Chieh Cheng, Ruihan Yang, Yang Fu, Linghao Chen, Jiarui Xu, Xinyu Zhang, Xinyue Wei, Jiteng Mu, and Jianglong Ye for their efforts in testing and evaluating the teleoperation system.

\bibliographystyle{plainnat}
\bibliography{references}

\newpage
\clearpage
\newpage
\input{sections/appendix.tex}
\end{document}

%% file: sections/intro.tex
\section{Introduction}
A grand goal of robotics is to endow robots with human-level intelligence to physically interact with the environment. Teleoperation~\cite{niemeyer2016telerobotics}, as a direct means to acquire human demonstrations for teaching robots, has been a powerful paradigm to approach this goal~\cite{kofman2007robot, du2012markerless, zhang2018deep, hedayati2018improving, mandlekar2020human, chen2022asha, jang2022bc, arunachalam:arxiv2022, muelling2015autonomy, shaw2022videodex, ye2023learning}. Compared to gripper-based manipulators, teleoperating dexterous hand-arm systems poses unprecedented challenges and often requires specialized apparatus that comes with high costs and setup efforts, such as Virtual Reality (VR) devices~\cite{holodex, hedayati2018improving, gharaybeh2019telerobotic}, wearable gloves~\cite{liu2017glove, liu2019high}, handheld controller~\cite{rakita2017motion, rakita2019remote, khadir2019teleoperator}, haptic sensors~\cite{elsner2022parti, kumar2015mujoco, salvato2022predicting, son2013human}, or motion capture trackers~\cite{zhao2012combining}. Fortunately, recent developments in vision-based teleoperation~\cite{antotsiou2018task, li2019vision, handa2020dexpilot, li2022dexterous, qin:ral2022, liang2020hand, kofman2005teleoperation, kofman2007robot, aronson2022gaze} have provided a low-cost and more generalizable alternative for teleoperating dexterous robot systems.

Despite the progress, the current paradigm of vision-based teleoperation systems still falls short when it comes to scaling up data collection for robot teaching. First, prior systems are often designed and engineered towards a particular robot model or deployment environment. For example, some systems rely on vision-based hand tracking models trained on datasets collected in the deployed studio~\cite{li2019vision, handa2020dexpilot}, and some rely on human-robot retargeting models~\cite{zhang2021human, fang2020vision} or collision avoidance models~\cite{sivakumar2022robotic} trained for the particular robot at use. These systems will scale poorly as the pool of robot models expands and the variety of operating environments increases. Second, each system is created and coupled with one specific ``reality'', either only in the real world or with a particular choice of simulators. For example, the HAPTIX~\cite{kumar2015mujoco} motion capture system is only developed for teleoperation in MuJoCo-based environments~\cite{todorov2012mujoco}. To facilitate large-scale data collection with simulation as well as closing sim-to-real gaps, we need teleoperation systems to operate both in virtual (with arbitrary choices of simulators) and in the real world. Finally, existing teleoperation systems are often tailored for single-operator and single-robot settings. To teach robots how to collaborate with other robot agents as well as with human agents, a teleoperation system should be designed to support multiple pilot-robot partners where the robots can physically interact with each other in a shared environment.

In this paper, we aim to set the foundation for scaling up data collection with vision-based dexterous teleoperation, by filling in the aforementioned gaps. To this end, we propose \textit{\oursnospace}, a unified and general teleoperation system (Fig.~\ref{fig:teaser}), which can be used for: 
\begin{itemize}
    \item \textit{Diverse} robot arm and dexterous hand models;
    \item \textit{Diverse} realities, i.e. different choices of simulators or the real world;
    \item Teleoperation from \textit{diverse} geographic locations, via a browser-based web visualizer developed for remote visual feedback;
    \item \textit{Diverse} camera configurations, e.g. RGB camera with or without depth, single or multiple cameras; 
    \item \textit{Diverse} operator-robot partnerships, e.g. two operators separately piloting two robots to collaboratively solve a manipulation task.
\end{itemize}

To achieve this goal, we first develop a general and high-performance motion retargeting library to translate human motion to robot motion in real time without learned models. 
Our collision avoidance module is also learning-free and powered by CUDA-based geometry queries. They can adapt to new robots given only the kinematic model, i.e., URDF files. Second, we develop a web-based viewer compatible with standard browsers, to achieve simulator-agnostic visualization and enable remote teleoperation across the internet. Third, we define a general software interface for visual-based teleoperation, which standardizes and decouples each module inside the teleoperation system. It enables smooth deployment on different simulators or real hardware. 

While being very general to support many settings with a single system, our system can still achieve great performance in the experiments. For real-world teleoperation, \ours can outperform a previous system~\cite{sivakumar2022robotic} designed for specific robot hardware \textbf{with higher success rates on 8 out of 10 tasks} proposed in their paper, using the same robot as ~\cite{sivakumar2022robotic}. For simulated environment teleportation, the smoother and collision-free demonstrations collected by \ours can bring better imitation learning results \textbf{with higher success rates on 5 out of 6 tasks} proposed in their paper, compared with a previous system~\cite{qin:ral2022} specifically designed for that simulator. Finally, we demonstrate that \ours can be extended to support collaborative manipulation, which to our best knowledge has neither been achieved in the literature of vision-based teleoperation nor on dexterous hands.

Our system is also packaged to be easily-deployable. The containerized design makes installation easy and frees users from handling software dependencies. We are committed to open-sourcing the system and benefiting the community.

%% file: sections/related.tex
\section{Related Work}

\begin{figure*}[!t]
 \centering
 \includegraphics[width=0.96\linewidth]{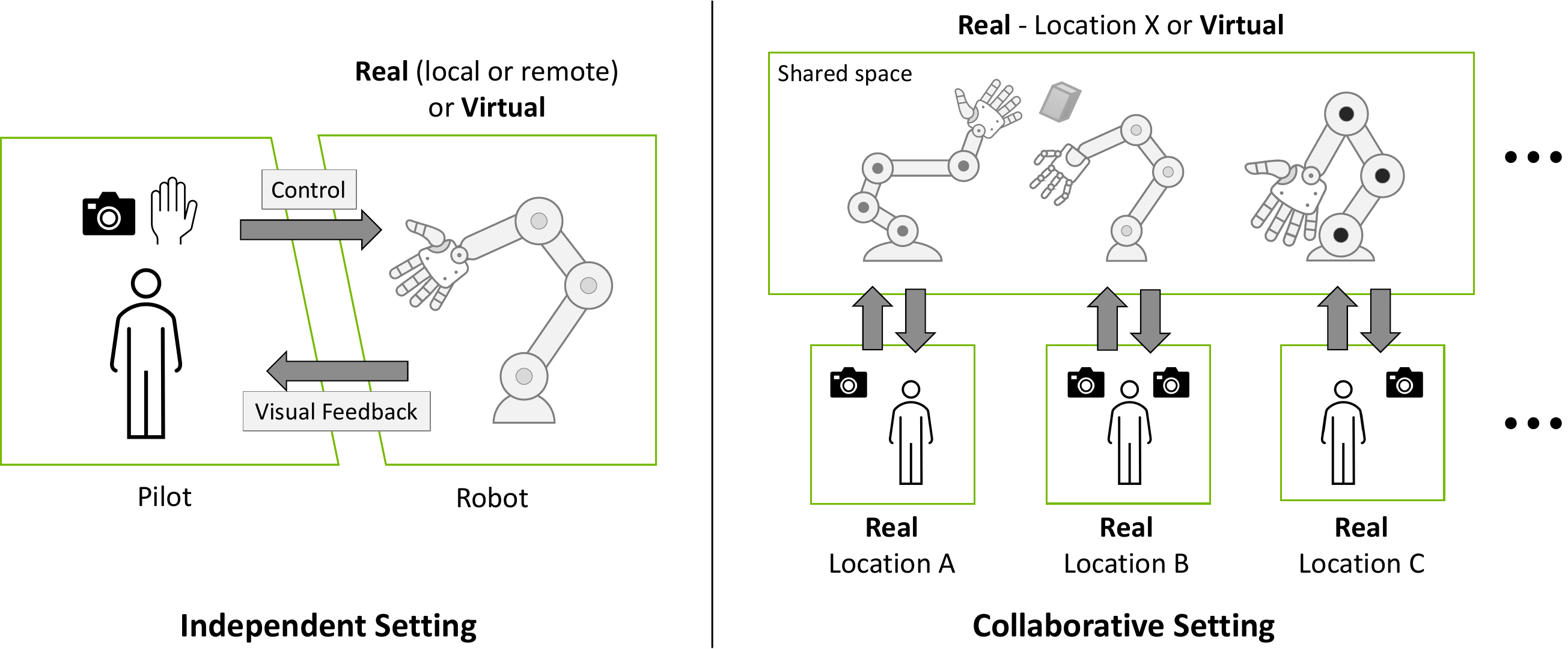}
 \caption{Paradigms of vision-based teleoperation systems in \textit{independent} and \textit{collaborative} settings. The system should support any arm-hand models, existed in either virtual or the real world, can operate with flexible camera configurations, provide visual feedback for both local or remote presence, and support multiple robots piloted in a shared space.}
 \label{fig:paradigm}
\end{figure*}

\textbf{Vision-based Robot Teleoperation.} 
Recent years have witnessed an increasing interest in teleoperation of dexterous robot hands by human hands. 
It relies on accurate tracking of human hand motions and finger articulations. 
Compared to the costly wearable hand tracking solutions, such as gloves~\cite{liu2017glove, liu2019high, mosbach:humanoids2022}, marker-based motion capture systems~\cite{zhao2012combining, liu2021semi}, inertia sensors~\cite{zhang2018feasibility} or VR headsets~\cite{holodex, lipton2017baxter, rosen2020mixed, ponomareva2021grasplook}, vision-based hand tracking is particularly favorable due to its low cost and low intrusion to the human operator. 
Early research in vision-based teleoperation focused on improving the performance of hand tracking~\cite{theobalt2004pitching, wang2009real, ajili2017gesture} and mapping human hand pose to robot hand pose~\cite{li2019vision, antotsiou2018task, arunachalam:arxiv2022, meeker2020continuous}. Recent works have expanded the scope of teleoperating a single robot hand to complete arm-hand systems~\cite{handa2020dexpilot, li:iros2020, zhang2018deep}.
However, these systems are designed and engineered towards a particular robot model (e.g., Kuka arm with Allegro hand in~\cite{handa2020dexpilot}, and PR2 arm with Shadow hand in ~\cite{li2022dexterous}), and rely on retargeting or collision detection models trained for specific robot hardware (e.g., Allegro hand and XArm6)~\cite{handa2020dexpilot, sivakumar2022robotic}
, making them difficult to transfer to new arm-hand systems and new environments. 
In contrast, our system is highly modularized with a versatile hand-tracking solution compatible with an arbitrary number of cameras, and configurable robot hand retargeting and motion generation modules for easy adaption to various robot arms and robot hand choices. 
This allows our system to achieve better performance compared to prior systems on various tasks while generalizing to a set of robot arm-hand systems and multiple environments.

\textbf{Teleoperation in Different Reality.} 
Manipulation with a dexterous robot hand is challenging due to its high degree of freedom. 
In recent years, dexterous robot teleoperation has been actively studied and shown promising progress in controlling a multi-fingered hand
to perform manipulation tasks in the real world~\cite{holodex, handa2020dexpilot, mizera2019evaluation}  by leveraging the morphological similarity between the dexterous hand and the human hand. 

With the advancement of data-driven approaches for robot manipulation~\cite{elsner2022parti, huang2021novel}, there is a growing need to collect human demonstrations in robotics. 
To enable easy and scalable data collection, teleoperation has also gained attention in simulated environments~\cite{xiang2020sapien, isaacgym, todorov2012mujoco, gan2020threedworld, coumans2016pybullet}. This provides a scalable solution to data collection by eliminating the need for real hardware, while maintaining access to oracle world information. 
For example, Mandlekar et al.~\cite{mandlekar2018roboturk} developed a crowd-sourcing platform to teleoperate robots via mobile devices as controllers. 
Tung et al.~\cite{tung2021learning} further extended this framework to allow multi-arm collaborative teleoperation. 
The above frameworks rely on inertial sensors for control signals and thus are limited to parallel-gripper and simple tasks such as pick-and-place. 
Our system offers the ability to perform a wide range of dexterous tasks with robots of different morphologies by utilizing state-of-the-art techniques in perception, optimization, and control. 
In addition, \ours is designed to support teleoperation in both virtual and the real world with a unified framework.

%% file: sections/overview.tex
\input{sections/big_table.tex}

\section{System Overview}
\label{sec:overview}

Fig.~\ref{fig:paradigm} illustrates our proposed paradigms of vision-based teleoperation systems. Below we introduce the features and designs of our system which realize the paradigms.

\subsection{System Features}
\label{sec:features}

\begin{enumerate}
    \item \textbf{Any arm-hand.} As shown in Fig.~\ref{fig:teaser}, \ours is designed for arbitrary dexterous arm-hand systems that are not limited to any specific robot type. 
    \item \textbf{Any reality.} \ours is decoupled from specific hardware drivers or physics simulators. It can support different realities as visualized in Fig.~\ref{fig:teaser}.
    \item \textbf{Anywhere remote teleoperation.} \ours provides a web-based visualizer to monitor the teleoperation and simulation in standard web browsers, e.g. Chrome. 
    \item \textbf{Any camera configuration.} \ours can consume data from both RGB and RGB-D cameras, and from either single or multiple cameras. Most importantly, it does not require extrinsic calibration as in most previous systems. This allows more flexible camera configurations and lower deployment overhead.
    \item \textbf{Any number of operator-robot partnerships}. \ours supports collaborative settings where operators separately pilot two robots to collaboratively solve a manipulation task.
    \item \textbf{Simple deployment.} \ours and all libraries are encapsulated as a docker image that can be downloaded and deployed on any Linux machine, which frees users from handling troublesome dependencies.
\end{enumerate}

Table~\ref{tab:main-comparision} compares \ours with other vision-based dexterous teleoperation systems. We compare the systems in three dimensions: (i) sensor requirements; (ii) robot-related support; (iii) afforded use cases. Among all these teleoperation systems,  \ours is the only one which can support different robot arms and enable collaborative teleoperation. It is also one of the only two systems that can support different dexterous hands.

\subsection{System Design}
\begin{figure*}[!t]
    \vspace{-1em}
    \centering
    \includegraphics[width=\linewidth]{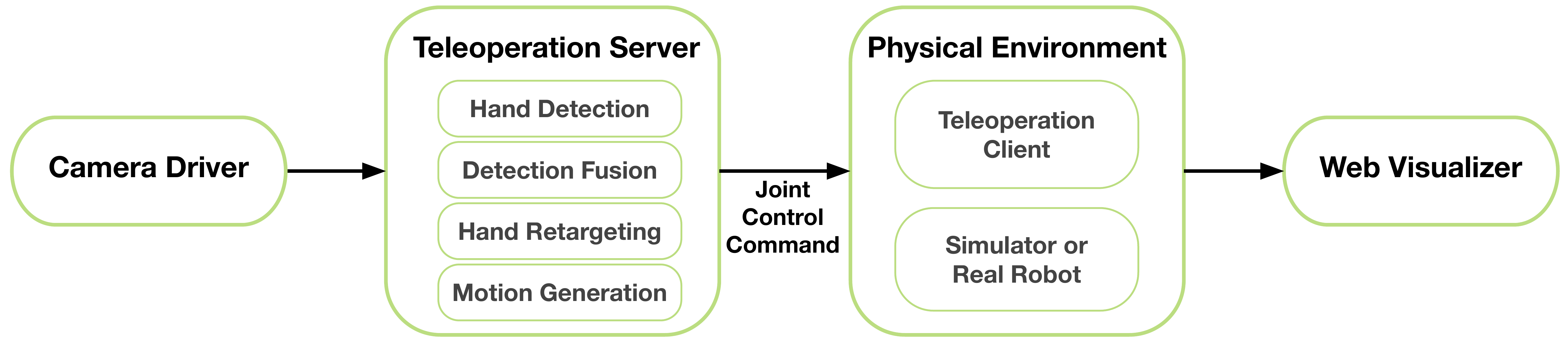}
    \caption{\textbf{System Architecture.} \ours is composed of four components: (i) camera driver, which captures the human hand pose in RGB or RGB-D format; (ii) teleportation server, the core component in our system, which performs hand pose detection and converts detection results to robot control commands; (iii) teleoperated robot, which is either a real robot or a simulated robot in a virtual environment; (iv) web visualizer, which enables remote visualization across the internet.} 
    \label{fig:overview}
    \vspace{-1.5em}
\end{figure*}

The architecture of the teleoperation system is shown in Fig.~\ref{fig:overview}. The teleoperation server (Section~\ref{sec:server}) receives the camera stream from the driver, detects the hand pose, and then converts it to joint control commands. The client receives these commands via network communication and uses them to control a simulated or real robot.  The system is designed with three key principles: modularity, communication-focused, and containerization. Modularity is achieved by implementing well-defined input-output interfaces for each sub-component, allowing for wide applicability to different robot arms, dexterous hands, cameras, and realities. Communication-focused design allows for remote and collaborative teleoperation and reduces computation requirements on the operator's side by deploying heavy computations on a powerful server. Finally, the containerized design makes installation and deployment easier compared to other robotics systems with heavy software dependencies.

%% file: sections/big_table.tex
\begin{table*}[!t]
    \centering
    \begin{tabular}{cccccccccc}
        \toprule%
        & \multicolumn{3}{c}{{{\bfseries Sensor Requirements}}} &
        \multicolumn{4}{c}{{{\bfseries Robot-related Support}}} &
        \multicolumn{2}{c}{{{\bfseries Use Case}}} \\

        \cmidrule[0.4pt](lr{0.125em}){2-4}%
        \cmidrule[0.4pt](lr{0.125em}){5-8}%
        \cmidrule[0.4pt](l{0.125em}){9-10}%
         
        & Calibration & Contact & \multirow{1}{*}{Depth}
        & Multiple & Multiple & \multirow{2}{*}{Reality} & Collision
        & Remote & Collaborative \\

        & Free & Free & Free & Arms & Hands && Free & Teleop & Teleop \\

        \cmidrule[0.4pt](r{0.125em}){1-10}

        DexPilot~\cite{handa2020dexpilot} & \no & \yes & \no & \no & \no & Real & \yes & \no & \no \\

        \myrowcolour
        Holo-Dex~\cite{holodex} & \yes & \yes & \yes & No Arm & \no & Real & \no & \yes & \no \\

        DIME~\cite{arunachalam:arxiv2022} & \no & \yes & \yes & No Arm & \no & Real & \no & \yes & \no \\

        \myrowcolour
        TeachNet~\cite{li2019vision} & \yes & \yes & \no & No Arm & \no & Sim\&Real & \no & \no & \no \\

        Telekinesis~\cite{sivakumar2022robotic} & \yes & \no & \yes & \no & \no & Real & \yes & \yes & \no \\

        \myrowcolour
        Qin \etal~\cite{qin:ral2022} & \yes & \yes & \no & No Arm & \yes & Sim & \no & \yes & \no \\

        MVP-Real~\cite{mvpreal} & \no & \no & \yes & No Arm & \no & Real & \no & \yes & \no \\

        \myrowcolour
        Transteleop~\cite{li2022dexterous} & \no & \no & \no & \no & \no & Real & \yes & \yes & \no \\

        Mosbach \etal~\cite{mosbach:humanoids2022} & \no & \no & \yes & \no & \no & Sim & \no & \yes & \no \\

        \myrowcolour
        \highest{\ours} & \yes & \yes & \yes & \yes & \yes & Sim\&Real & \yes & \yes & \yes \\

        \bottomrule
    \end{tabular}
    \caption{\textbf{Comparison of Vision-Based Teleoperation System.} We compare \oursnospace 's capabilities with related visual teleoperation system for multi-fingered dexterous robots. ``No Arm'' in the column of ``Multiple Arms'' means this system can only control hand motion but not arm-hand systems.}
    \label{tab:main-comparision}

    \vspace{-2em}
\end{table*}

%% file: sections/teleop_system.tex
\section{Teleoperation Server}
\label{sec:server}
The teleoperation server, outlined in Section~\ref{sec:overview}, utilizes the RGB or RGB-D data from one or multiple cameras and generates smooth and collision-free control commands for the robot arm and dexterous hand. It consists of four modules: (i) the hand pose detection module, which predicts hand wrist and finger poses from the camera stream, (ii) the detection fusion module, which integrates the results from multiple cameras, (iii) the hand pose retargeting module, which maps human hand poses to the dexterous robot hand, and (iv) the motion generation module, which produces high-frequency control signals for the robot arm. A standardized software interface is defined for all four modules to facilitate flexibility and generalizability in \ours.

\subsection{Hand Pose Detection}
The hand pose detection module offers a unique feature to utilize input from various camera configurations, including RGB or RGB-D cameras, and single or multiple cameras. The design principle is to leverage more information, such as depth, and additional cameras, to improve performance when available. But it can also perform the task with minimal input, i.e. a single RGB camera. The detection module has two outputs: local finger keypoint positions in the wrist frame and global 6D wrist pose in the camera frame. The finger keypoint detection only requires RGB data while the wrist pose detection can optionally use depth information to achieve better results.

\textbf{Finger Keypoint Detection.}
Our finger keypoint detection utilizes MediaPipe~\cite{zhang2020mediapipe}, a lightweight, RGB-based hand detection tool that can operate in real-time on a CPU. The MediaPipe detector can accurately locate 3D keypoints of 21 hand-knuckle coordinates in the wrist frame and 2D keypoints on the image. 

\textbf{Wrist Pose Detection from RGB-D.}
We use the pixel positions of the detected keypoints to retrieve the corresponding depth values from the depth image. Then, utilizing known intrinsic camera parameters, we compute the 3D positions of the keypoints in the camera frame. The alignment of the RGB and depth images is handled by the camera driver. With the 3D keypoint positions in both the local wrist frame and global camera frame, we can estimate the wrist pose using the Perspective-n-Point (PnP) algorithm.

\textbf{Wrist Pose Detection from RGB only.}
The orientation of the hand can be computed analytically from the local positions of the detected keypoints. However, determining the wrist position in the camera frame can be challenging without explicit 3D information. To enhance MediaPipe for global wrist pose estimation, we adopt the approach used in FrankMocap~\cite{rong2020frankmocap} by incorporating an additional neural network that predicts the weak perspective transformation scale of the hand. The weak perspective transformation approximates the original perspective camera model by assuming that the observed object is farther from the camera than its size. Together with intrinsic parameters, this scale factor can be used to approximate the 3D position of the hand. The wrist position computed this way has a larger error than depth camera, but it is still sufficient for many downstream teleoperation tasks.

\subsection{Detection Fusion}
The detection fusion module integrates multiple camera detection results. Self-occlusion can be a problem when performing hand pose detection, especially when the hand is perpendicular to the camera plane. Using multiple cameras can alleviate this problem by providing additional views. However, there are two main challenges in fusing multiple detection results: (i) each camera can only estimate the hand pose in its own frame and (ii) there is no straightforward metric to quantify the confidence of each detection result.

To overcome the first challenge, we perform an auto-calibration process using the human hand as a natural marker. We use the first $N$ frames of hand detection results from multiple cameras to calculate the relative rotation between each camera, expressed in $SO(3)$. We find that although the absolute position of detected hand pose is not so accurate in RGB-only setting, the relative motion between consecutive frames is more robust. With orientation between each camera, we can transform the detected relative motion from different cameras into a single frame.

To address the second challenge, we use the SMPL-X~\cite{SMPL-X:2019} hand shape parameters predicted from the detection module, as inspired by Qin \etal~\cite{qin:ral2022}. During teleoperation, the true shape parameters should remain constant for a given operator, but the predicted values can contain errors during self-occlusion. We observe that larger shape parameter prediction errors often correspond to larger pose errors. To approximate the confidence score, we take the mean of the estimated shape parameters in the first $N$ frames as a reference and compute the error between the predicted shape parameters and the reference. 
Implementation-wise, we require the operator to spread their fingers during the first $N$ frames to ensure an accurate reference value of shape parameters. The fusion module then selects the relative motion captured by the camera with the highest confidence score and forwards it to the next module. In implementation, we choose $N=50$.

\begin{figure*}[!t]
    \vspace{-1em}
    \centering
    \includegraphics[width=\linewidth]{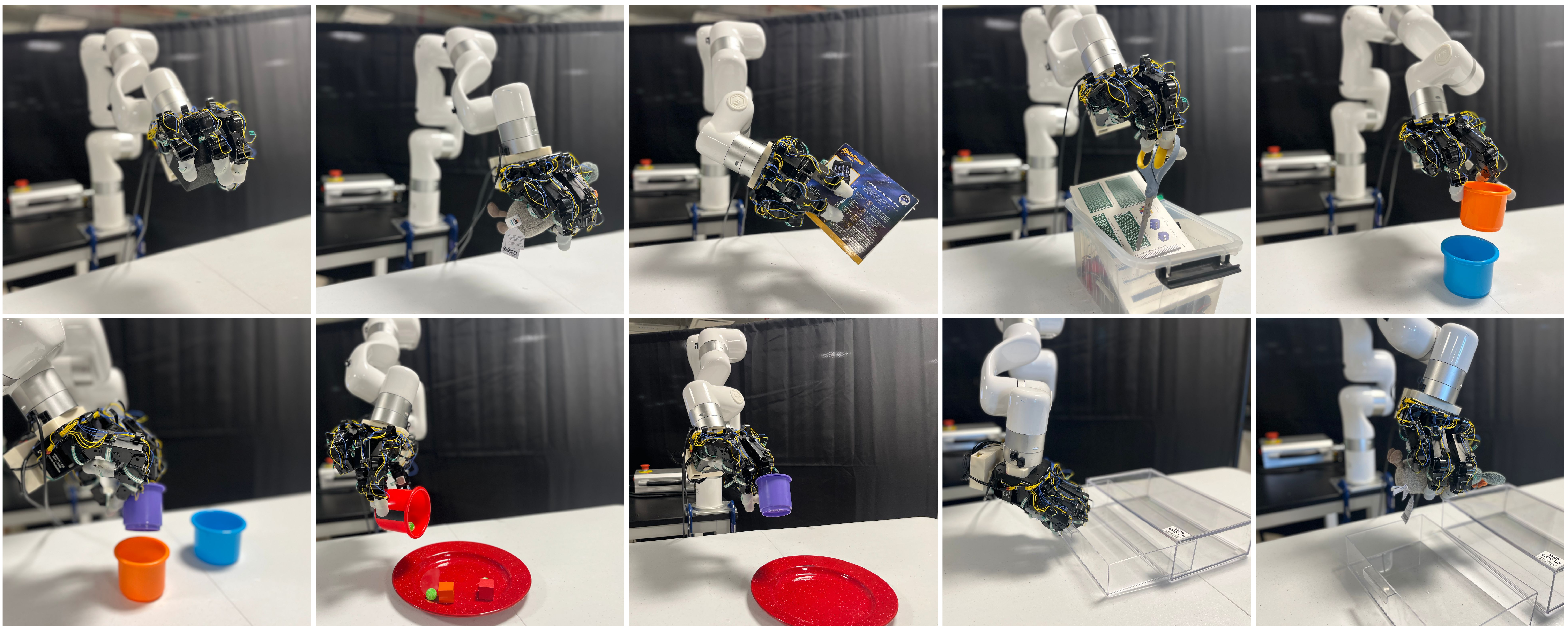}
    \caption{\textbf{Real Robot Teleoperation Tasks.} We replicate the ten manipulation tasks proposed in Sivakumar \etal~\cite{sivakumar2022robotic} using same or similar objects. Top row, left to right: Pickup Box Object, Pickup Fabric Toy, Box Rotation, Scissor Pickup, Cup Stack. Bottom row, left to right: Two Cup Stacking, Pouring Cubes onto Plate, Cup Into Plate, Open Drawer and Open Drawer and Pickup Cup. }
    \label{fig:realworld-task}
    \vspace{-2em}
\end{figure*}

\subsection{Hand Pose Retargeting}
The hand pose retargeting module maps the human hand pose data obtained from perception algorithms into joint positions of the teleoperated robot hand. This process is often formulated as an optimization problem~\cite{qin2021dexmv, handa2020dexpilot}, where the difference between the keypoint vectors of the human and robot hand is minimized. 
The optimization can be defined as follows:

\begin{equation}
\begin{split}
\min_{q_t} \sum_{i=0}^{N} || \alpha v_t^i - f_i(q_t)& ||^2 + \beta||q_{t} - q_{t-1}||^2 \\
\mathrm{s.t.} \quad q_{l} \le q_t & \le q_{u},
\end{split}
\end{equation}
where $q_t$ represents the joint positions of the robot hand at time step $t$, $v^i_t$ is the $i$-th keypoint vector for human hand computed from the detected finger keypoints, $f_i(q_t)$ is the $i$-th forward kinematics function which takes the robot hand joint positions $q_t$ as input and computes the $i$-th keypoint vector for the robot hand, $q_l$ and $q_u$ are the lower and upper limits of the joint position, $\alpha$ is a scaling factor to account for hand size difference. An additional penalty term with weight $\beta$ is included to improve temporal smoothness. When retargeting to a different morphology, such as a Dclaw in Figure~\ref{fig:teaser}, we need to specify the keypoint vectors mapping between the robot and human fingers manually.
It is worth noting that this module only considers the robot hand. 

\subsection{Motion Generation}

Given the detected wrist and hand pose,  our goal is to generate smooth and collision-free motion of robot arm to reach the target Cartesian end-effector pose. 
Real-time motion generation methods are required to have a smooth teleoperation experience. 
In the prior work of~\cite{handa2020dexpilot}, the robot motion is driven by Riemannian Motion Policies (RMPs)~\cite{rmp,rmp_flow} that can calculate acceleration fields in real-time. 
However, accelerations towards a particular end-effector pose do not guarantee natural trajectories. 
In this work, we adopt CuRobo~\cite{Sundaralingam2023CuRobo}, a highly parallelized collision-free robot motion generation library accelerated by GPUs, to generate natural and reactive robot motion in real-time. 
In \oursnospace, the motion generation module receives the Cartesian pose of the end-effector at a low frequency ($25$ Hz) from the hand detection and retargeting modules, and generates collision-free joint-space trajectories within joint limits at a higher frequency ($120$ Hz). 
The generated trajectories are ready for safe execution by impedance controllers on either a simulated or real robot. 

%% file: sections/web_visualization.tex
\section{Web-based Teleoperation Viewer}
\label{sec:web-viz}

To better support the teleoperation tasks, we implement a web-based visualization module to facilitate remote and collaborative teleoperation, especially for teleoperation in simulated environments. It has the following features: (i) browser-based viewer, which makes it easily accessible remotely; (ii) synchronized 
 visualization, i.e. two operators working on the same collaborative task should see the same scene synchronously from their own local view ports. The viewer is developed based upon the \texttt{meshcat}~\cite{meshcat} library and utilize \texttt{Three.js}~\cite{danchilla2012three} for rendering. The visualization server ports the simulation results onto the browser after each simulation iteration. Operators can get visual feedback from the browser window and move their hands to control the corresponding robot. More details about the implementation of our viewer can be found in the supplementary materials.

%% file: sections/system_eval.tex
\section{System Evaluation}
\label{sec:system_eval}

\begin{table}[t]
    \centering
    \begin{tabular}{cccc}
       \toprule%
       \multicolumn{1}{c}{\multirow{3}{*}{\bfseries HardWare}}  &  &  \multicolumn{1}{c}{Desktop} & \multicolumn{1}{c}{Laptop} \\

       & GPU  &  \multicolumn{1}{c}{RTX 3090} & \multicolumn{1}{c}{RTX 2070} \\

       & CPU & \multicolumn{1}{c}{i9-10980XE} & \multicolumn{1}{c}{i7-8750} \\

       
       \cmidrule[0.4pt](r{0.125em}){1-4} 

       \multicolumn{1}{c}{\multirow{6}{*}{\bfseries Profiling}} & Modules &  Time (ms) & Time (ms) \\

       &\CC Hand Pose (RGB) &\CC $26\pm5$  &\CC $34\pm5$ \\

       & Hand Pose (RGB-D) & $27\pm5$    & $35\pm5$ \\

       &\CC Fusion &\CC ~$1\pm0$ &\CC ~$1\pm0$ \\

       & Retargeting &  ~$9\pm7$   & $10\pm9$ \\

       &\CC Motion &\CC ~$8\pm3$  &\CC $11\pm5$ \\
       
       \bottomrule
       
    \end{tabular}
    \caption{\textbf{Profiling Results.} We profile different modules inside teleoperation server on both desktop and laptop. The time is measured when all teleoperation modules are run on the same computer simultaneously.}
    \label{tab:profiling}
    \vspace{-2em}
\end{table}

\begin{table*}[!t]
        \begin{minipage}{0.4\linewidth}    
	\centering
	\includegraphics[width=\linewidth]{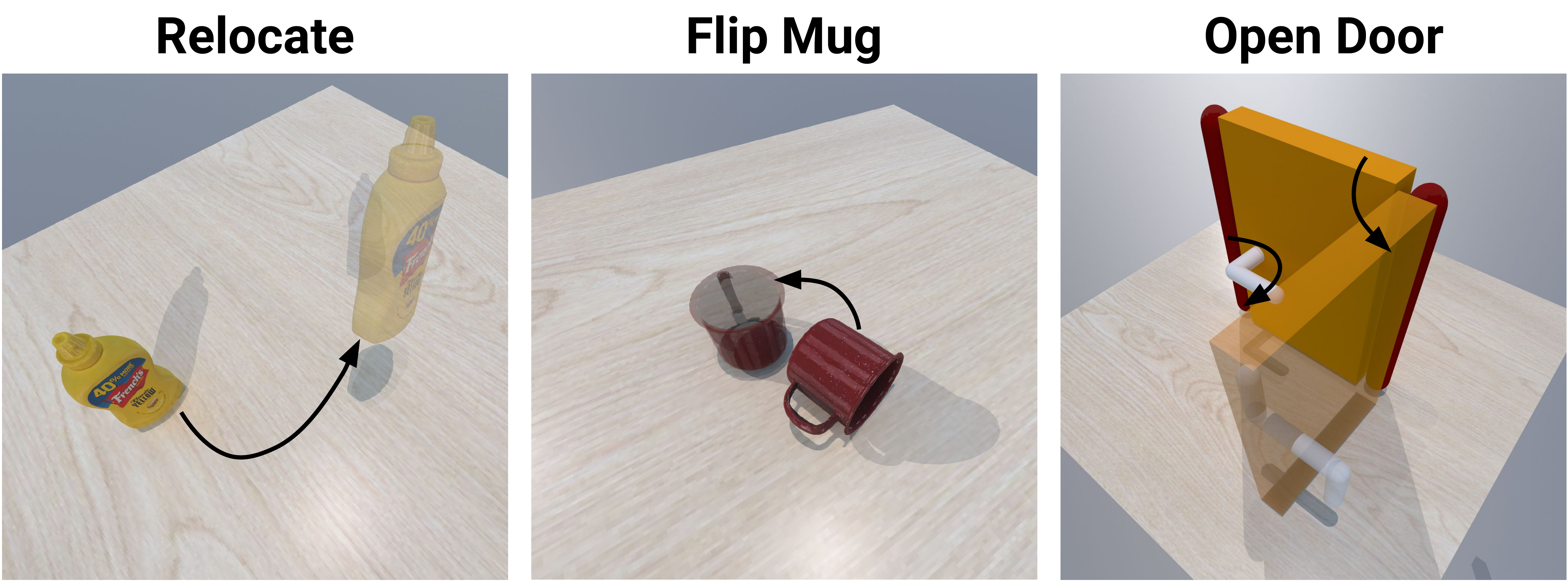}
	\end{minipage}
	\begin{minipage}{0.60\linewidth}
	\centering
        \begin{tabular}{ccccc}
        \toprule
        \multicolumn{2}{c}{Manipulation Task}  & RL & Baseline~\cite{qin:ral2022} & Ours \\
        \cmidrule[0.4pt](r{0.125em}){1-5} 
        \multirow{3}{*}{Floating-Hand} 
        & Relocate & $36.3\pm15.3$ & $49.7\pm18.3$ & $\textbf{53.7}\pm\textbf{12.2}$ \\
        & \CC Flip Mug & \CC $33.7\pm15.0 $ &\CC $\textbf{51.3}\pm\textbf{34.7}$ &\CC $47.3\pm28.3$ \\
        & Open Door & $69.3\pm38.0$ & $64.7\pm14.7$ & $\textbf{73.3}\pm~\textbf{9.0}$ \\
        \multirow{3}{*}{Arm-Hand} 
        &\CC Relocate &\CC $33.7\pm29.3$   &\CC $40.3\pm36.7$ &\CC $\textbf{70.0}\pm~\textbf{9.8}$ \\
        & Flip Mug & $31.0\pm28.7$  & $36.0\pm32.4$ & $\textbf{53.7}\pm\textbf{24.0}$ \\
        &\CC Open Door &\CC $34.7\pm31.7$  &\CC $51.3\pm30.7$ &\CC $\textbf{79.7}\pm\textbf{15.5}$ \\
        \bottomrule
        \end{tabular}
	\end{minipage}
 
	\captionof{table}{\textbf{Imitation Learning Experiments in SAPIEN Environments.} The left figure visualizes the three tasks we use for both teleoperation data collection and imitation learning. The transparent object represents the goal of the task while the black arrow represents the steps of the task. The right table shows the success rate of the evaluated methods. We use $\pm$ to represent the mean and standard deviation over three random seeds. The success rate is computed from 100 trials. }
    \label{tab:sapien_imitation}
    \hspace{-2em}
\end{table*}

\subsection{Profiling Analysis}
\label{sec:profiling}

We perform profiling on modules mentioned in Section~\ref{sec:server} on a desktop and a laptop. As shown in Table~\ref{tab:profiling}, the most time-consuming module is hand pose detection, which runs on a GPU for real-time inference. The designed maximum frequency for hand pose detection is 25Hz, so both the desktop and laptop can meet the requirement. Both the retargeting module and the fusion module run at the same frequency as the hand detection module due to the publisher and subscriber logic. For best performance, the motion generation module should run at 120Hz but can still work with a lower frequency. Notably, we found it difficult to achieve this throughput when running all these modules on the same computer. Luckily, with our communication-oriented design, we can run the control modules on a separate machine to achieve the best performance.

\begin{table}[t]
    \centering
    \begin{tabular}{ccc}
       \toprule
       Task &  \ours & Telekinesis~\cite{sivakumar2022robotic} \\
       \cmidrule[0.4pt](r{0.125em}){1-3} 
       Pickup Box Object & \textbf{1.0} & 0.9\\ \myrowcolour
       Pickup Fabric Toy & \textbf{1.0} & 0.9\\
       Box Rotation & \textbf{0.6} & \textbf{0.6}\\ \myrowcolour
       Scissor Pickup & \textbf{0.8} & 0.7 \\
       Cup Stack &  \textbf{0.9} & 0.6 \\ \myrowcolour
       Two Cup Stacking & \textbf{0.7} & 0.3 \\
       Pouring Cubes onto Plate & \textbf{0.7} & \textbf{0.7} \\ \myrowcolour
       Cup Into Plate & \textbf{1.0} & 0.8 \\
       Open Drawer & \textbf{1.0}  & 0.9 \\ \myrowcolour
       Open Drawer and Pickup Object & \textbf{0.9} & 0.6 \\    
       \bottomrule
    \end{tabular}
    \caption{\textbf{Real Robot Teleoperation Results.} We replicate the experiment settings and tasks in \cite{sivakumar2022robotic} and compare with \cite{sivakumar2022robotic}. For the baseline method, we use the success rate reported in their paper~\cite{sivakumar2022robotic}}
    \label{tab:real_teleop_comp}
\end{table}

\subsection{Real Robot Teleoperation}
\label{sec:real_robot_exp}

In this section, we will test our \ours system across a wide range of real-world tasks that covers diverse objects and manipulation skills. Besides, we will compare our teleoperation performance of \ours with a similar teleoperation system. A fair comparison of real-robot tasks is often very challenging due to the difficulty in replicating the baseline methods delicately. To ensure a more fair comparison, we replicate the ten manipulation tasks proposed in Robotic Telekinesis~\cite{sivakumar2022robotic} with the same XArm6 robot, Allegro hand, and similar objects. A trained operator attempt to solve this tasks using \ours system. The ten tasks are visualized in Fig.~\ref{fig:realworld-task}. Same as ~\cite{sivakumar2022robotic}, we run each task ten times for \ours and use a single Intel RealSense camera. For the baseline method, we directly use the results reported in their paper. 

As shown in Table~\ref{tab:real_teleop_comp}, \ours can get a higher success rate of 8/10 tasks and the same success rate on 2/10 compared with the baseline. Although \ours is designed to be more general, it can still outperform the baseline system that was specifically designed for the XArm6-Allegro hardware. We find that the major advantage of our system is the capability to handle objects with thin-walled structures, such as the cup-stack, two-cup-stacking, and cup-into-plate tasks. Our optimization-based retargeting module can close the distance between finger tips, which makes grasping the cup more stable. However, the network-based retargeting can hardly translate the fine-grained precision grasp from human to robot, which leads to a lower success rate.

%% file: sections/application.tex
\section{Applications}

\subsection{Imitation Learning}
\label{sec:imitation_learning}

The most important application of the proposed system is imitation learning from demonstration. We can first collect demonstrations on several dexterous manipulation tasks and then use the data to train imitation learning algorithms. In this experiment, we will show that the teleoperation data collected using our \ours can better support downstream imitation learning tasks. In the following subsection, we will first introduce the experiment setting and baseline and then discuss the experimental results. 

\textbf{Baseline and Comparison.}
To fairly compare with previous teleoperation systems, we need to align both the task setting and robot configuration precisely. It is often challenging for real-robot hardware but much easier for a simulated environment. Thus, we choose a recent vision-based teleoperation work~\cite{qin:ral2022} that can be used for simulated robots as our baseline. It is worth noting that we are comparing two teleoperation systems via the demonstration data collected by each system. Thus, we compared with the baseline by training the same learning algorithm on different demonstration data collected via the baseline system and our teleoperation system. We follow \cite{qin:ral2022} to choose Demo Augmented Policy Gradient (DAPG)~\cite{Rajeswaran2018} as the imitation algorithm. We also compare it with a pure reinforcement learning (RL) based algorithm from \cite{qin2023dexpoint} which does not utilize demonstrations. We provide the same dense reward for RL training as previous work~\cite{qin:ral2022}.

\begin{figure}[!t]
    \vspace{-1em}
    \centering
    \includegraphics[width=\linewidth]{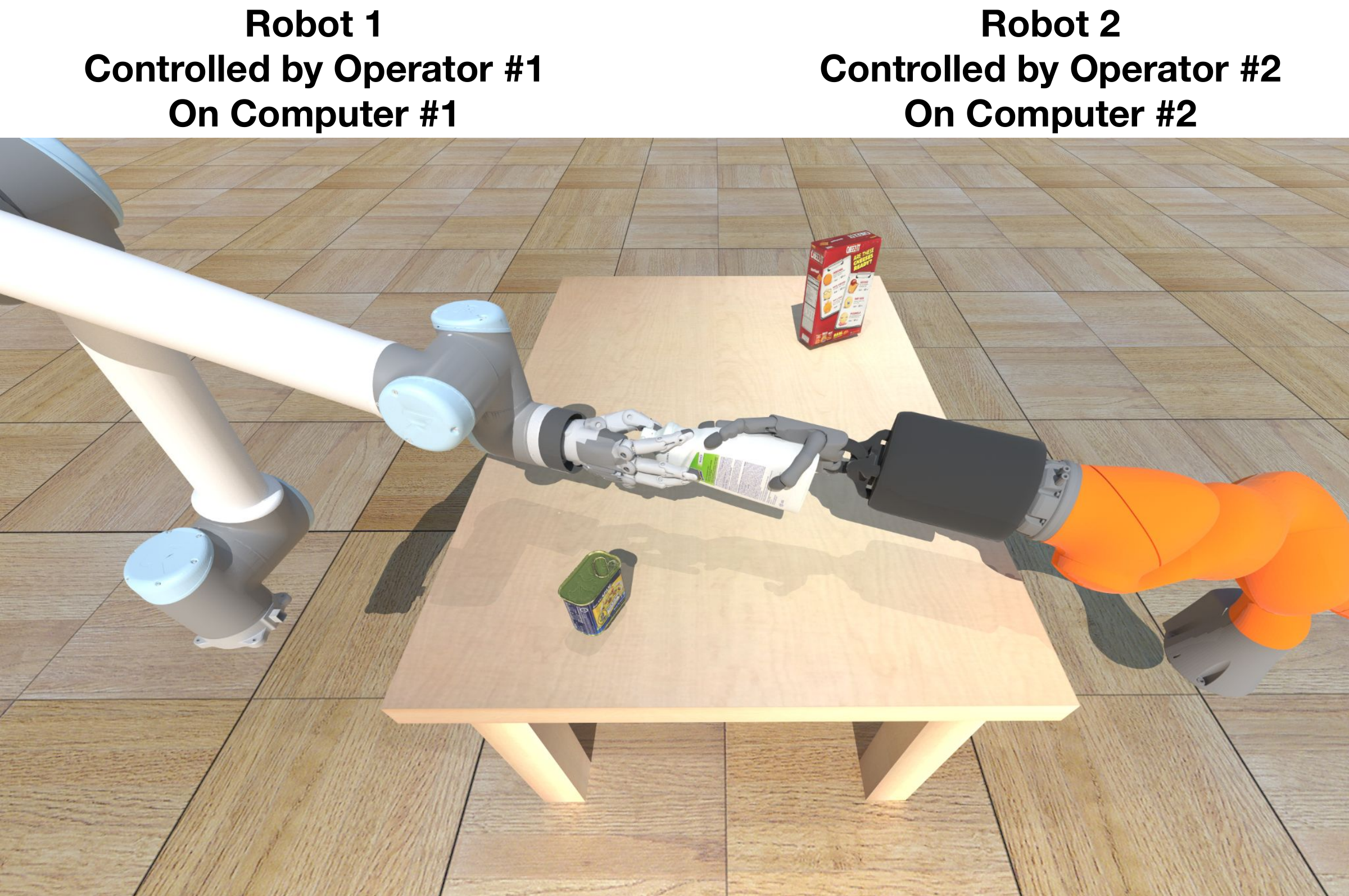}
    \caption{\textbf{Collaborative Teleoperation for Handover Task.} Operator $\#1$ act as the UR10-Schunk robot and operator $\#2$ acts as the Kuka-Shadow robot. In this task, the operator $\#2$ needs to pick up an object on the table and handover it to operator $\#1$.}
    \label{fig:handover-task}
    \vspace{-2em}
\end{figure}

\textbf{Manipulation Tasks.}
We directly use the manipulation tasks proposed by the baseline work~\cite{qin:ral2022} for comparison, which include three tasks: (i) \textit{Relocate}, where the robot picks an object on the table and moves it to the target position; (ii) \textit{Flip Mug}, where the robot needs to rotate the mug for 90 degrees to flip it back; (iii) \textit{Open Door}, where the robot needs to first rotate the lever to unlock the door, and then pull it to open the door. These tasks are visualized in the left figure of Table~\ref{tab:sapien_imitation}. The manipulated objects in all three tasks are randomly initialized and the target position is also randomized in \emph{Relocate}.
Each manipulation task has two variants: the floating-hand variant and the arm-hand variant. The floating-hand is a dexterous hand without a robot arm that can move freely in space. The arm-hand means the hand is mounted on a robot arm with a fixed base, which is a more realistic setting. 

\textbf{Demonstration Details.}
For the baseline teleoperation system~\cite{qin:ral2022}, we directly use the demonstration collected by the original authors with 50 demonstration trajectories for each task. The baseline system only utilizes a single RGB-D camera. For fairness, we also collect 50 trajectories for each task using the single camera setup. The baseline system can only handle floating hands and they propose a demonstration translation pipeline to convert the demonstration with floating hands to demonstrations with arm-hand. For our \ours, we collect demonstrations using the arm-hand setting and convert the demonstration to floating hand so that the demonstration can be used by both the floating-hand variant and arm-hand variant.

\begin{figure}[!t]
    \vspace{-1em}
    \centering
    \includegraphics[width=\linewidth]{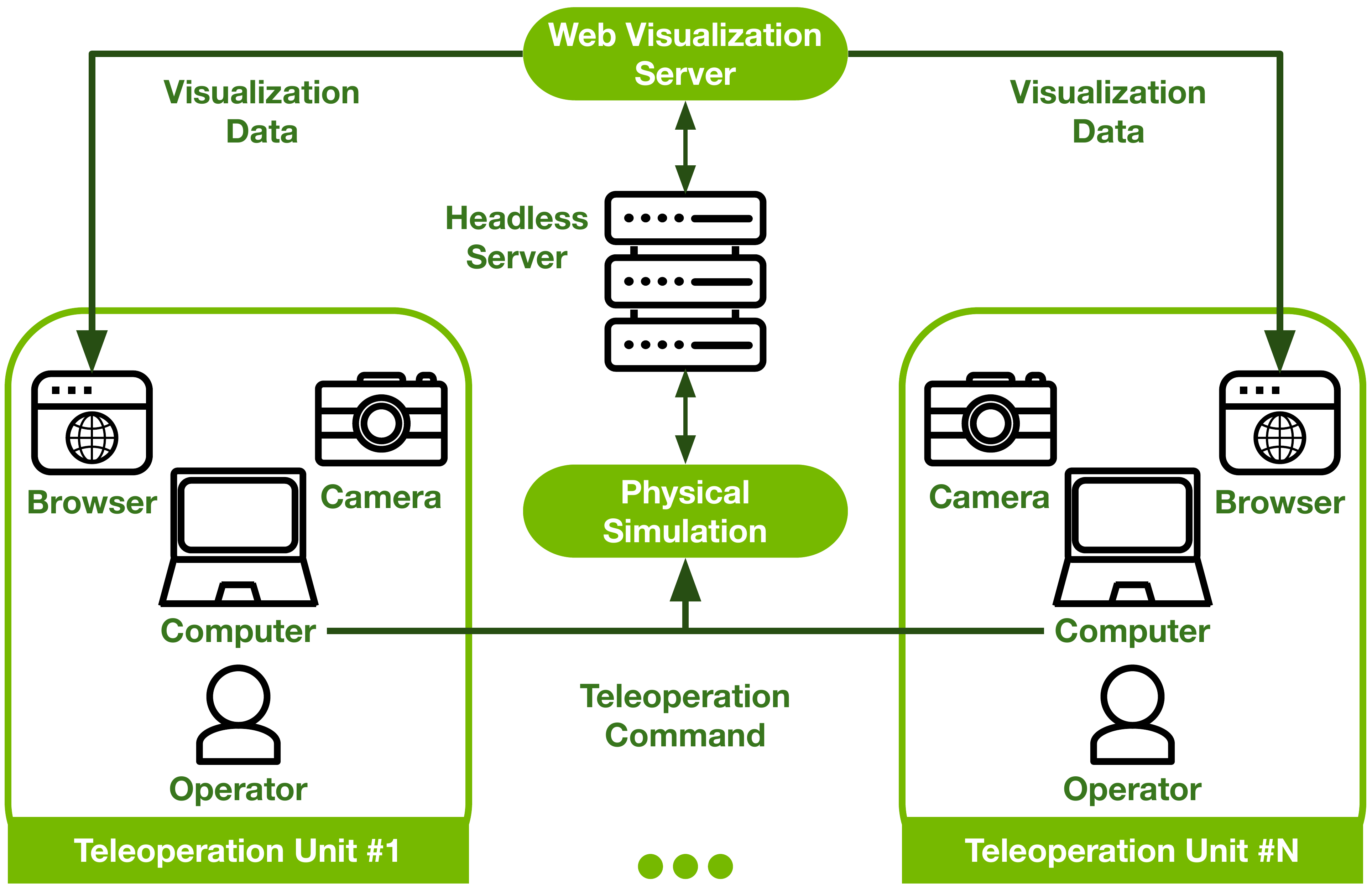}
    \caption{\textbf{Collaborative Teleoperation System.} Our system can be extended to collaborative manipulation tasks even when operators are not in the same physical location. Each operator can use a local computer with camera to detect the hand pose and send the detection results to a central server. Meanwhile, they can use the web browser to visualize the current simulation environment, including the robot controlled by other operators. }
    \label{fig:collaborative_system}
    \vspace{-2em}
\end{figure}

\textbf{Results and Discussion.}
For each method on each task, we train policies with three different random seeds. For each policy, we evaluate it on 100 trials. More details about the success metrics can be found in~\cite{qin:ral2022}.  As shown in Table~\ref{tab:sapien_imitation}, the imitation learning algorithm trained on demonstration collected by \ours can outperform baseline and RL on most tasks with one exception.
Compared with the demonstration collected via the baseline system, our system has two benefits that contribute to better performance in imitation learning: (i) The collected trajectory is more smooth, which means that the state-action pairs are more consistent and easier to be consumed by the network. (ii) Different from the baseline, our system explicitly supports teleoperation with arm-hand system and guarantees no self-collision. On the contrary, the baseline system utilizes retargeting to generate joint trajectory for robot arm, which may lead to several self-collision for robot arm. Thus we can observe significant performance gain of our system for manipulation tasks with arm-hand.
For the flip mug task, the difficulty of collecting demonstration with arm is much larger than with a floating hand, which influences the demonstration quality.

\subsection{Collaborative Manipulation}
\label{sec:collaborative_teleop}

Collaborative manipulation is a key technology for the development of human-robot systems~\cite{tung2021learning}. Collecting demonstration data for collaborative manipulation tasks has been a challenging task since it requires multiple operators to work together seamlessly. With our modularized and extensible system design and web-based visualization, our system enables convenient data collection on collaborative tasks, even if operators are not in the same physical location. In this section, we show that our teleoperation system can be extended to a collaborative setting where multiple operators coordinate together to perform manipulation tasks. We choose human-to-robot handover as an example as shown in Fig.~\ref{fig:handover-task}. In this setting, operator $\#1$ control a robot hand, and operator $\#2$ control a human hand.  

\textbf{Collaborative Teleoperation System Design.} Fig.~\ref{fig:collaborative_system} illustrates the system architecture for multi-operator collaboration, which includes two components. (i) Teleoperation Units: It is composed of a computer that is connected to at least one camera and a human operator. In each teleoperation unit, the human operator will watch the real-time visualization on a web browser and move the hand accordingly to perform manipulation tasks. 
(ii) Central Server: it runs the physical simulation and the web visualization server. The detection results from multiple teleoperation units are sent to the server and converted into robot control commands based on the pipeline in Section~\ref{sec:server}. Meanwhile, the web visualizer server will keep synchronized with the simulated environment and maintain the visualization resources as described in Section~\ref{sec:web-viz}.

%% file: sections/appendix.tex
\appendix

\subsection{Supplementary Overview}
This supplementary material provides more details, results and visualizations accompanying the main paper, including
\begin{itemize}
    \item More details and visualization about \textbf{teleoperation server}, including detection and retargeting modules;
    \item More details about \textbf{web-based teleoperation viewer};
    \item Additional experimental results on \textbf{system evaluation}.
\end{itemize}

More visualization can be found at our project page: {\color{blue}{\texttt{\url{http://anyteleop.com}}}}.

\subsection{Teleoperation Server}
In this section, we will show more intermediate results from our system, including visualization of hand pose detection results and the retargeting results of various robot hands.

\begin{figure*}[!t]
\vspace{-1em}
    \centering
    \includegraphics[width=\linewidth]{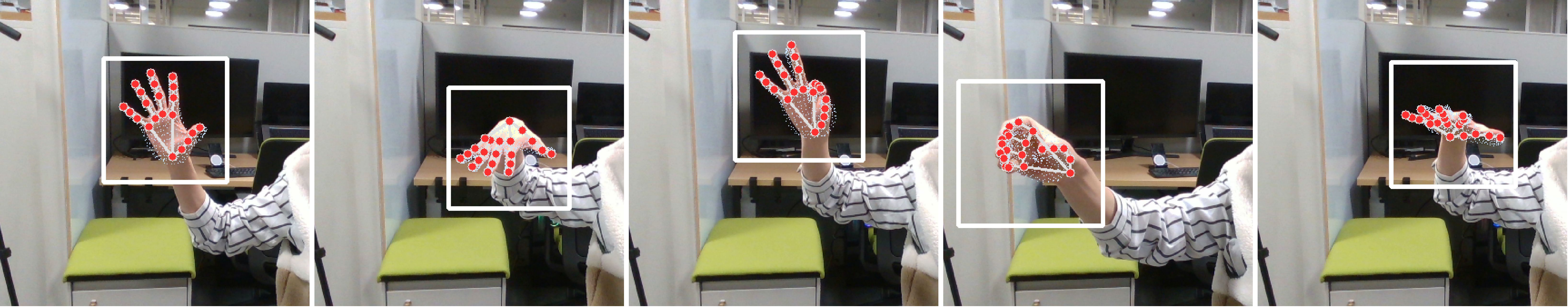}
    \caption{\textbf{Hand Pose Detection Visualization.} This figure visualizes the hand detection results, with the white bounding box highlighting the predicted area and red points marking the identified finger key points. The hand skeleton is represented by the grey lines connecting the key points. Additionally, the small grey points depict the 2D projection of 3D vertices from the SMPL-X hand model. The figure showcases five diverse cases, from left to right: (i) a hand spreading out the fingers to initiate teleoperation, (ii) fingers facing downwards in preparation for a top-down grasp, (iii) a precision grasp using the thumb and index finger, (iv) a power grasp using all five fingers, and (v) a failure scenario where the hand is positioned vertically relative to the camera plane.} 
    \label{fig:hand_pose}
\end{figure*}

\textbf{Visualization of Hand Pose Detection}
We visualize the hand pose detection results in Figure~\ref{fig:hand_pose}. We showcase five typical cases, which include: (i) a hand spreading out the fingers for teleoperation initialization, (ii) fingers facing downwards in preparation for a top-down grasp, (iii) a precision grasp using the thumb and index finger, (iv) a power grasp using all five fingers, and (v) a failure case where the hand is positioned vertically relative to the camera plane.

\textbf{Visualization of Hand Pose Retargeting}
We demonstrate the results of hand pose retargeting in Figure~\ref{fig:retargeting}. The figure displays seven gestures being performed using four different dexterous hands.

\subsection{Web-based Teleoperation Viewer}
\begin{figure*}[!t]
\vspace{-1em}
    \centering
    \includegraphics[width=\linewidth]{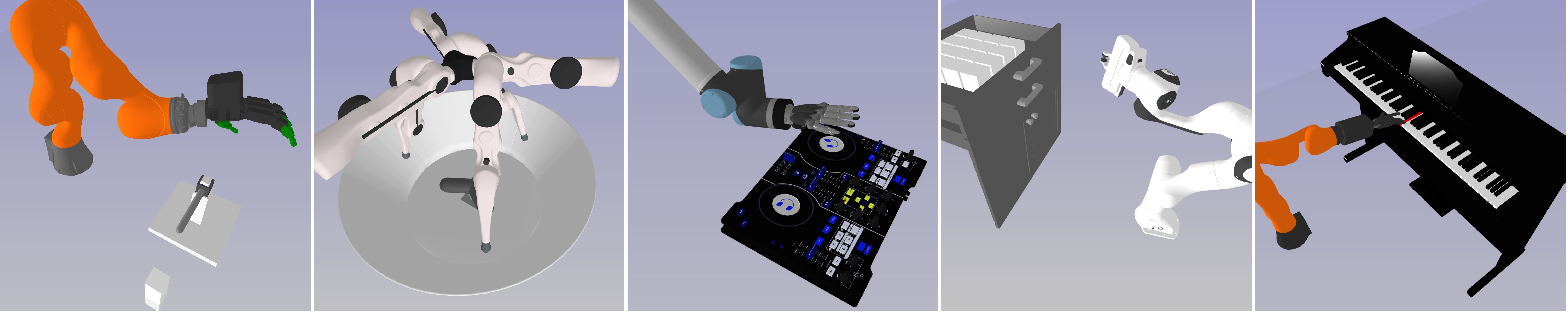}
    \caption{\textbf{Web-based Viewer.} The visualization of teleoperation process can be perform in the web-based viewer, which features the five tasks from the IsaacGym tasks shown in the Figure 1 of the main paper. The viewer utilizes the \textit{three.js} library for real-time rendering through a web browser.} 
    \label{fig:web_viewer}
\end{figure*}

In this section, we demonstrate how the web-based visualizer provides accessibility and multi-view support for teleoperation through its lightweight rendering and capability to run in multiple browser windows. Figure~\ref{fig:web_viewer} shows screenshots of the web-based visualizer when it is used to visualize the five IsaacGym tasks depicted in the Figure 1 in the main paper. 

\textbf{Lightweight Rendering vs High Visual Quality.} 
The design of our web-based viewer prioritizes accessibility and convenience, as it can be used on any device with a browser and provides minimal but sufficient rendering capabilities for teleoperation. Although the rendering quality may not be as advanced as the original simulator viewer, simulation states can be saved for offline rendering to produce high-quality visual data. For example, in visual reinforcement learning tasks using RGB images as inputs, the rendered data can be generated using a more powerful engine such as a ray tracer after teleoperation is completed.

\begin{figure*}[!t]
    \centering
    \includegraphics[width=0.95\linewidth]{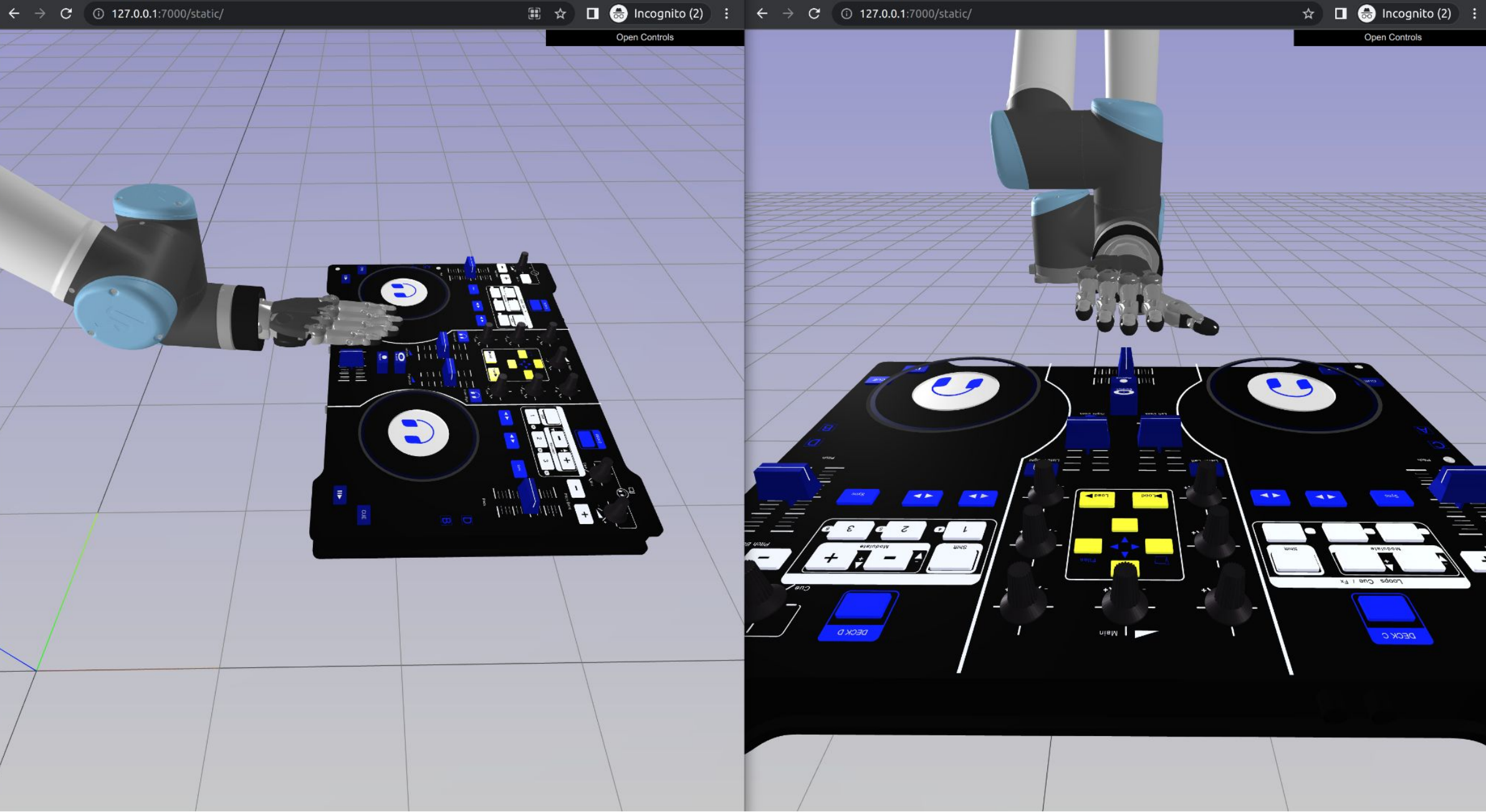}
    \caption{\textbf{Multi-view Support with More Browser Windows.} The web-based viewer offers multiple views to enhance the operator's understanding of the 3D object relationships. Additional views can be accessed by simply opening more browser windows.} 
    \label{fig:multi_view}
\end{figure*}

\begin{figure*}[!t]
\vspace{-1em}
    \centering
    \includegraphics[width=0.75\linewidth]{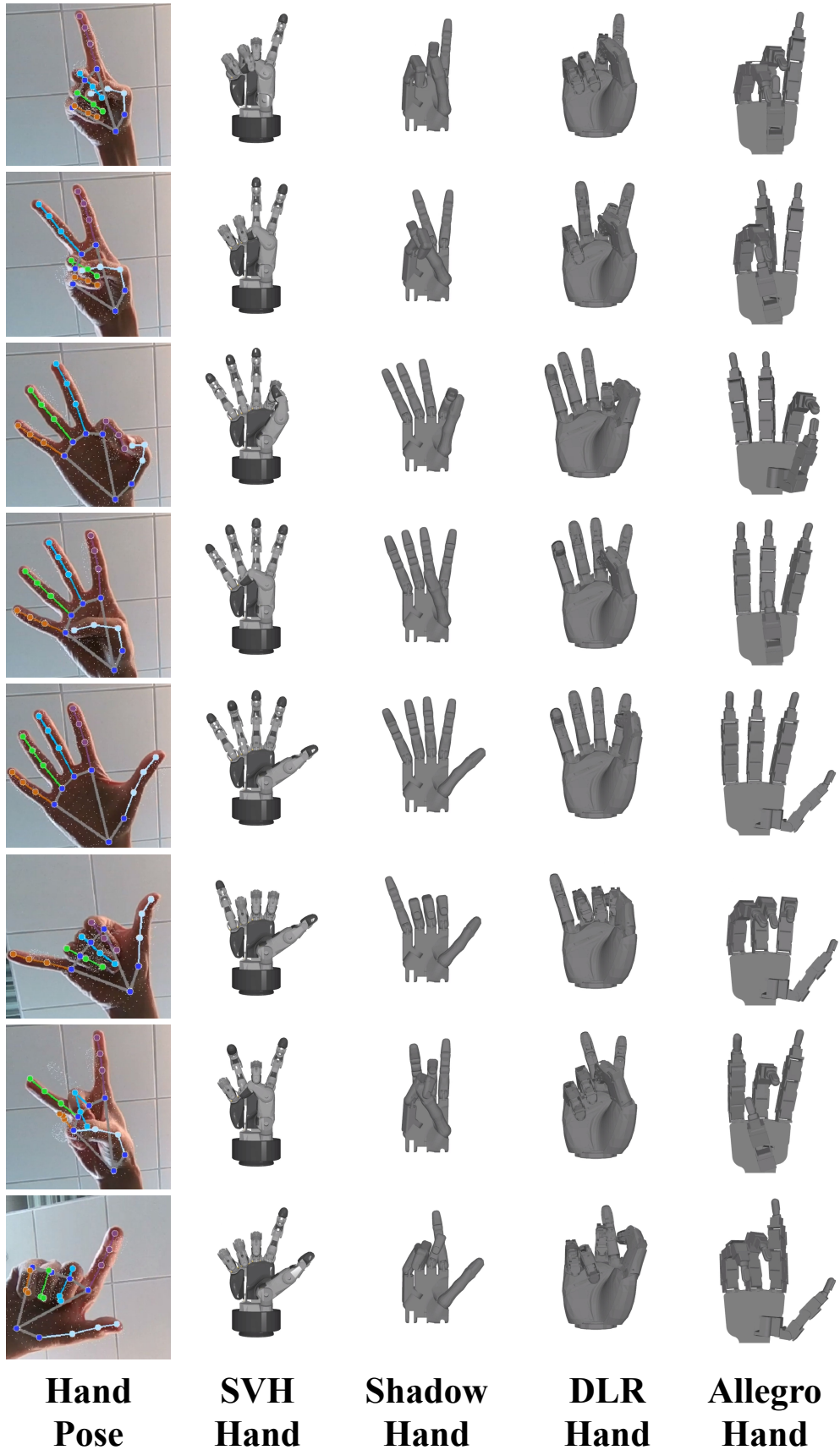}
    \caption{\textbf{Visualization of Hand Pose Retargeting.} The figure presents the results of hand pose retargeting for seven gestures and four different dexterous robot hands. The four hands are displayed in order from left to right: (i) Schunk SVH hand; (ii) Shadow Hand; (iii) DLR Hand; (iv) Allegro Hand.} 
    \label{fig:retargeting}
\end{figure*}

\textbf{Multi-View Support for Teleoperation.} In teleoperation, human operators often require a clear understanding of the spatial relationships between objects and robots to make informed decisions. This information can be provided through multi-view rendering, which is a widely used technique in previous teleoperation works~\cite{li2022dexterous, qin:ral2022, wei2021multi}. Our web-based viewer offers multi-view support to the operator by simply opening multiple browser windows. As shown in Figure~\ref{fig:multi_view}, an example of the operator using two views to perform a manipulation task is displayed. The operator is able to open as many windows as needed to enhance their teleoperation experience.

\section{System Evaluation on Camera Configurations}

\begin{table}[t]
    \centering
    \begin{tabular}{ccc}
       \toprule
       Camera Configuration &  Completion Time & Error Percentage (\%) \\
       \cmidrule[0.4pt](r{0.125em}){1-3} 
       Single RGB & $109s$ & $28.1\%$ \\ 
       Single RGB-D & $87s$ & $21.8\%$\\ 
       Two RGB-D & \textbf{$\mathbf{74}$s} & \textbf{$\mathbf{12.5\%}$} \\ 
       \bottomrule
    \end{tabular}
    \caption{\textbf{Comparison of Camera Configurations.} We evaluate the teleoperation performance on the Play Piano task with different camera configurations. }
    \label{tab:camera_config}
\end{table}

In this section, we examine the impact of different camera configurations on the teleoperation performance of our system, \ours, which is capable of supporting diverse configurations including RGB, RGB-D, and single or multiple cameras. Even with a minimal configuration, i.e. a single RGB camera, the system can still perform effectively. Additionally, by adding more resources, such as multiple cameras, our system can achieve better performance.

We use the \textit{Play Piano} task implemented in IsaacGym~\cite{isaacgym} as the evaluation scenario, which requires the robot hand to press piano keys in a specific order. The task is shown in the Figure 1 in the main paper and the video in the supplementary material. To quantify performance, we introduce two task metrics: (i) completion time, i.e. the elapsed time from start to finish, and (ii) the percentage of incorrect key presses, which measures the number of incorrectly pressed keys relative to the total number of keys.

A trained operator performs the task ten times for each camera configuration. As reported in Table~\ref{tab:camera_config}, 
with additional information, such as depth, and increasing number of cameras, the task can be completed faster and with fewer errors, which demonstrates that our system allows users to easily trade-off between efficiency and system cost based on their use case.

%% file: paper_template.bbl
\begin{thebibliography}{68}
\providecommand{\natexlab}[1]{#1}
\providecommand{\url}[1]{\texttt{#1}}
\expandafter\ifx\csname urlstyle\endcsname\relax
  \providecommand{\doi}[1]{doi: #1}\else
  \providecommand{\doi}{doi: \begingroup \urlstyle{rm}\Url}\fi

\bibitem[Ajili et~al.(2017)Ajili, Mallem, and Didier]{ajili2017gesture}
Insaf Ajili, Malik Mallem, and Jean-Yves Didier.
\newblock Gesture recognition for humanoid robot teleoperation.
\newblock In \emph{2017 26Th IEEE international symposium on robot and human
  interactive communication (RO-MAN)}, pages 1115--1120. IEEE, 2017.

\bibitem[Antotsiou et~al.(2018)Antotsiou, Garcia-Hernando, and
  Kim]{antotsiou2018task}
Dafni Antotsiou, Guillermo Garcia-Hernando, and Tae-Kyun Kim.
\newblock Task-oriented hand motion retargeting for dexterous manipulation
  imitation.
\newblock In \emph{ECCV Workshops}, 2018.

\bibitem[Aronson and Admoni(2022)]{aronson2022gaze}
Reuben~M Aronson and Henny Admoni.
\newblock Gaze complements control input for goal prediction during assisted
  teleoperation.
\newblock In \emph{Robotics science and systems}, 2022.

\bibitem[Arunachalam et~al.(2022{\natexlab{a}})Arunachalam, G{\"u}zey,
  Chintala, and Pinto]{holodex}
Sridhar~Pandian Arunachalam, Irmak G{\"u}zey, Soumith Chintala, and Lerrel
  Pinto.
\newblock Holo-dex: Teaching dexterity with immersive mixed reality.
\newblock \emph{arXiv preprint arXiv:2210.06463}, 2022{\natexlab{a}}.

\bibitem[Arunachalam et~al.(2022{\natexlab{b}})Arunachalam, Silwal, Evans, and
  Pinto]{arunachalam:arxiv2022}
Sridhar~Pandian Arunachalam, Sneha Silwal, Ben Evans, and Lerrel Pinto.
\newblock Dexterous imitation made easy: A learning-based framework for
  efficient dexterous manipulation.
\newblock \emph{arXiv preprint arXiv:2203.13251}, 2022{\natexlab{b}}.

\bibitem[Chen et~al.(2022)Chen, Gao, Reddy, Berseth, Dragan, and
  Levine]{chen2022asha}
Sean Chen, Jensen Gao, Siddharth Reddy, Glen Berseth, Anca~D Dragan, and Sergey
  Levine.
\newblock Asha: Assistive teleoperation via human-in-the-loop reinforcement
  learning.
\newblock In \emph{2022 International Conference on Robotics and Automation
  (ICRA)}, pages 7505--7512. IEEE, 2022.

\bibitem[Cheng et~al.(2020)Cheng, Mukadam, Issac, Birchfield, Fox, Boots, and
  Ratliff]{rmp_flow}
Ching-An Cheng, Mustafa Mukadam, Jan Issac, Stan Birchfield, Dieter Fox, Byron
  Boots, and Nathan Ratliff.
\newblock Rmp flow: A computational graph for automatic motion policy
  generation.
\newblock In \emph{Algorithmic Foundations of Robotics XIII: Proceedings of the
  13th Workshop on the Algorithmic Foundations of Robotics 13}, pages 441--457.
  Springer, 2020.

\bibitem[Coumans and Bai(2016)]{coumans2016pybullet}
Erwin Coumans and Yunfei Bai.
\newblock {PyBullet}, a python module for physics simulation for games,
  robotics and machine learning.
\newblock \emph{GitHub repository}, 2016.

\bibitem[Danchilla and Danchilla(2012)]{danchilla2012three}
Brian Danchilla and Brian Danchilla.
\newblock Three. js framework.
\newblock \emph{Beginning WebGL for HTML5}, pages 173--203, 2012.

\bibitem[Deits(2018)]{meshcat}
Robin Deits.
\newblock Meshcat.
\newblock \url{https://github.com/rdeits/meshcat}, 2018.

\bibitem[Du et~al.(2012)Du, Zhang, Mai, and Li]{du2012markerless}
Guanglong Du, Ping Zhang, Jianhua Mai, and Zeling Li.
\newblock Markerless kinect-based hand tracking for robot teleoperation.
\newblock \emph{International Journal of Advanced Robotic Systems}, 9\penalty0
  (2):\penalty0 36, 2012.

\bibitem[Elsner et~al.(2022)Elsner, Reinerth, Figueredo, Naceri, Walter, and
  Haddadin]{elsner2022parti}
Jean Elsner, Gerhard Reinerth, Luis Figueredo, Abdeldjallil Naceri, Ulrich
  Walter, and Sami Haddadin.
\newblock Parti-a haptic virtual reality control station for model-mediated
  robotic applications.
\newblock \emph{Frontiers in Virtual Reality}, 3, 2022.

\bibitem[Fang et~al.(2020)Fang, Ma, Wang, and Sun]{fang2020vision}
Bin Fang, Xiao Ma, Jiachun Wang, and Fuchun Sun.
\newblock Vision-based posture-consistent teleoperation of robotic arm using
  multi-stage deep neural network.
\newblock \emph{Robotics and Autonomous Systems}, 131:\penalty0 103592, 2020.

\bibitem[Gan et~al.(2020)Gan, Schwartz, Alter, Mrowca, Schrimpf, Traer,
  De~Freitas, Kubilius, Bhandwaldar, Haber, et~al.]{gan2020threedworld}
Chuang Gan, Jeremy Schwartz, Seth Alter, Damian Mrowca, Martin Schrimpf, James
  Traer, Julian De~Freitas, Jonas Kubilius, Abhishek Bhandwaldar, Nick Haber,
  et~al.
\newblock Threedworld: A platform for interactive multi-modal physical
  simulation.
\newblock \emph{arXiv preprint arXiv:2007.04954}, 2020.

\bibitem[Gharaybeh et~al.(2019)Gharaybeh, Chizeck, and
  Stewart]{gharaybeh2019telerobotic}
Zaid Gharaybeh, Howard Chizeck, and Andrew Stewart.
\newblock \emph{Telerobotic control in virtual reality}.
\newblock IEEE, 2019.

\bibitem[Handa et~al.(2020)Handa, Van~Wyk, Yang, Liang, Chao, Wan, Birchfield,
  Ratliff, and Fox]{handa2020dexpilot}
Ankur Handa, Karl Van~Wyk, Wei Yang, Jacky Liang, Yu-Wei Chao, Qian Wan, Stan
  Birchfield, Nathan Ratliff, and Dieter Fox.
\newblock Dexpilot: Vision-based teleoperation of dexterous robotic hand-arm
  system.
\newblock In \emph{ICRA}, 2020.

\bibitem[Hedayati et~al.(2018)Hedayati, Walker, and
  Szafir]{hedayati2018improving}
Hooman Hedayati, Michael Walker, and Daniel Szafir.
\newblock Improving collocated robot teleoperation with augmented reality.
\newblock In \emph{International Conference on Human-Robot Interaction}, 2018.

\bibitem[Huang et~al.(2021)Huang, Wang, Bai, Huang, Sun, Xiao, and
  Yeatman]{huang2021novel}
Zebin Huang, Ziwei Wang, Weibang Bai, Yanpei Huang, Lichao Sun, Bo~Xiao, and
  Eric~M Yeatman.
\newblock A novel training and collaboration integrated framework for
  human--agent teleoperation.
\newblock \emph{Sensors}, 21\penalty0 (24):\penalty0 8341, 2021.

\bibitem[Jang et~al.(2022)Jang, Irpan, Khansari, Kappler, Ebert, Lynch, Levine,
  and Finn]{jang2022bc}
Eric Jang, Alex Irpan, Mohi Khansari, Daniel Kappler, Frederik Ebert, Corey
  Lynch, Sergey Levine, and Chelsea Finn.
\newblock Bc-z: Zero-shot task generalization with robotic imitation learning.
\newblock In \emph{Conference on Robot Learning}, pages 991--1002. PMLR, 2022.

\bibitem[Khadir et~al.(2019)Khadir, Varley, and
  Sindhwani]{khadir2019teleoperator}
Bachir~El Khadir, Jake Varley, and Vikas Sindhwani.
\newblock Teleoperator imitation with continuous-time safety.
\newblock \emph{arXiv preprint arXiv:1905.09499}, 2019.

\bibitem[Kofman et~al.(2005)Kofman, Wu, Luu, and
  Verma]{kofman2005teleoperation}
Jonathan Kofman, Xianghai Wu, Timothy~J Luu, and Siddharth Verma.
\newblock Teleoperation of a robot manipulator using a vision-based human-robot
  interface.
\newblock \emph{IEEE transactions on industrial electronics}, 52\penalty0
  (5):\penalty0 1206--1219, 2005.

\bibitem[Kofman et~al.(2007)Kofman, Verma, and Wu]{kofman2007robot}
Jonathan Kofman, Siddharth Verma, and Xianghai Wu.
\newblock Robot-manipulator teleoperation by markerless vision-based hand-arm
  tracking.
\newblock \emph{International Journal of Optomechatronics}, 2007.

\bibitem[Kumar and Todorov(2015)]{kumar2015mujoco}
Vikash Kumar and Emanuel Todorov.
\newblock Mujoco haptix: A virtual reality system for hand manipulation.
\newblock In \emph{International Conference on Humanoid Robots (Humanoids)},
  2015.

\bibitem[Li et~al.(2019)Li, Ma, Liang, G{\"o}rner, Ruppel, Fang, Sun, and
  Zhang]{li2019vision}
Shuang Li, Xiaojian Ma, Hongzhuo Liang, Michael G{\"o}rner, Philipp Ruppel, Bin
  Fang, Fuchun Sun, and Jianwei Zhang.
\newblock Vision-based teleoperation of shadow dexterous hand using end-to-end
  deep neural network.
\newblock In \emph{ICRA}, 2019.

\bibitem[Li et~al.(2020)Li, Jiang, Ruppel, Liang, Ma, Hendrich, Sun, and
  Zhang]{li:iros2020}
Shuang Li, Jiaxi Jiang, Philipp Ruppel, Hongzhuo Liang, Xiaojian Ma, Norman
  Hendrich, Fuchun Sun, and Jianwei Zhang.
\newblock A mobile robot hand-arm teleoperation system by vision and {IMU}.
\newblock In \emph{IROS}, 2020.

\bibitem[Li et~al.(2022)Li, Hendrich, Liang, Ruppel, Zhang, and
  Zhang]{li2022dexterous}
Shuang Li, Norman Hendrich, Hongzhuo Liang, Philipp Ruppel, Changshui Zhang,
  and Jianwei Zhang.
\newblock A dexterous hand-arm teleoperation system based on hand pose
  estimation and active vision.
\newblock \emph{IEEE Transactions on Cybernetics}, 2022.

\bibitem[Liang et~al.(2020)Liang, Handa, Van~Wyk, Makoviychuk, Kroemer, and
  Fox]{liang2020hand}
Jacky Liang, Ankur Handa, Karl Van~Wyk, Viktor Makoviychuk, Oliver Kroemer, and
  Dieter Fox.
\newblock In-hand object pose tracking via contact feedback and gpu-accelerated
  robotic simulation.
\newblock In \emph{2020 IEEE International Conference on Robotics and
  Automation (ICRA)}, pages 6203--6209. IEEE, 2020.

\bibitem[Lipton et~al.(2017)Lipton, Fay, and Rus]{lipton2017baxter}
Jeffrey~I Lipton, Aidan~J Fay, and Daniela Rus.
\newblock Baxter's homunculus: Virtual reality spaces for teleoperation in
  manufacturing.
\newblock \emph{IEEE Robotics and Automation Letters}, 3\penalty0 (1):\penalty0
  179--186, 2017.

\bibitem[Liu et~al.(2017)Liu, Xie, Millar, Edmonds, Gao, Zhu, Santos, Rothrock,
  and Zhu]{liu2017glove}
Hangxin Liu, Xu~Xie, Matt Millar, Mark Edmonds, Feng Gao, Yixin Zhu, Veronica~J
  Santos, Brandon Rothrock, and Song-Chun Zhu.
\newblock A glove-based system for studying hand-object manipulation via joint
  pose and force sensing.
\newblock In \emph{2017 IEEE/RSJ International Conference on Intelligent Robots
  and Systems (IROS)}, pages 6617--6624. IEEE, 2017.

\bibitem[Liu et~al.(2019)Liu, Zhang, Xie, Zhu, Liu, Wang, and Zhu]{liu2019high}
Hangxin Liu, Zhenliang Zhang, Xu~Xie, Yixin Zhu, Yue Liu, Yongtian Wang, and
  Song-Chun Zhu.
\newblock High-fidelity grasping in virtual reality using a glove-based system.
\newblock In \emph{2019 international conference on robotics and automation
  (icra)}, pages 5180--5186. IEEE, 2019.

\bibitem[Liu et~al.(2021)Liu, Jiang, Xu, Liu, and Wang]{liu2021semi}
Shaowei Liu, Hanwen Jiang, Jiarui Xu, Sifei Liu, and Xiaolong Wang.
\newblock Semi-supervised 3d hand-object poses estimation with interactions in
  time.
\newblock In \emph{CVPR}, 2021.

\bibitem[Makoviychuk et~al.(2021)Makoviychuk, Wawrzyniak, Guo, Lu, Storey,
  Macklin, Hoeller, Rudin, Allshire, Handa, et~al.]{isaacgym}
Viktor Makoviychuk, Lukasz Wawrzyniak, Yunrong Guo, Michelle Lu, Kier Storey,
  Miles Macklin, David Hoeller, Nikita Rudin, Arthur Allshire, Ankur Handa,
  et~al.
\newblock Isaac gym: High performance gpu-based physics simulation for robot
  learning.
\newblock \emph{arXiv preprint arXiv:2108.10470}, 2021.

\bibitem[Mandlekar et~al.(2018)Mandlekar, Zhu, Garg, Booher, Spero, Tung, Gao,
  Emmons, Gupta, Orbay, et~al.]{mandlekar2018roboturk}
Ajay Mandlekar, Yuke Zhu, Animesh Garg, Jonathan Booher, Max Spero, Albert
  Tung, Julian Gao, John Emmons, Anchit Gupta, Emre Orbay, et~al.
\newblock Roboturk: A crowdsourcing platform for robotic skill learning through
  imitation.
\newblock In \emph{Conference on Robot Learning}, pages 879--893. PMLR, 2018.

\bibitem[Mandlekar et~al.(2020)Mandlekar, Xu, Mart{\'\i}n-Mart{\'\i}n, Zhu,
  Fei-Fei, and Savarese]{mandlekar2020human}
Ajay Mandlekar, Danfei Xu, Roberto Mart{\'\i}n-Mart{\'\i}n, Yuke Zhu,
  Li~Fei-Fei, and Silvio Savarese.
\newblock Human-in-the-loop imitation learning using remote teleoperation.
\newblock \emph{arXiv preprint arXiv:2012.06733}, 2020.

\bibitem[Meeker et~al.(2020)Meeker, Haas-Heger, and
  Ciocarlie]{meeker2020continuous}
Cassie Meeker, Maximilian Haas-Heger, and Matei Ciocarlie.
\newblock A continuous teleoperation subspace with empirical and algorithmic
  mapping algorithms for nonanthropomorphic hands.
\newblock \emph{IEEE Transactions on Automation Science and Engineering},
  19\penalty0 (1):\penalty0 373--386, 2020.

\bibitem[Mizera et~al.(2019)Mizera, Delrieu, Weistroffer, Andriot, Decatoire,
  and Gazeau]{mizera2019evaluation}
C~Mizera, T~Delrieu, V~Weistroffer, C~Andriot, A~Decatoire, and J-P Gazeau.
\newblock Evaluation of hand-tracking systems in teleoperation and virtual
  dexterous manipulation.
\newblock \emph{IEEE Sensors Journal}, 20\penalty0 (3):\penalty0 1642--1655,
  2019.

\bibitem[Mosbach et~al.(2022)Mosbach, Moraw, and Behnke]{mosbach:humanoids2022}
Malte Mosbach, Kara Moraw, and Sven Behnke.
\newblock Accelerating interactive human-like manipulation learning with
  {GPU}-based simulation and high-quality demonstrations.
\newblock In \emph{Humanoids}, 2022.

\bibitem[Muelling et~al.(2015)Muelling, Venkatraman, Valois, Downey, Weiss,
  Javdani, Hebert, Schwartz, Collinger, and Bagnell]{muelling2015autonomy}
Katharina Muelling, Arun Venkatraman, Jean-Sebastien Valois, John Downey,
  Jeffrey Weiss, Shervin Javdani, Martial Hebert, Andrew~B Schwartz, Jennifer~L
  Collinger, and J~Andrew Bagnell.
\newblock Autonomy infused teleoperation with application to bci manipulation.
\newblock \emph{arXiv preprint arXiv:1503.05451}, 2015.

\bibitem[Niemeyer et~al.(2016)Niemeyer, Preusche, Stramigioli, and
  Lee]{niemeyer2016telerobotics}
G{\"u}nter Niemeyer, Carsten Preusche, Stefano Stramigioli, and Dongjun Lee.
\newblock Telerobotics.
\newblock \emph{Springer handbook of robotics}, pages 1085--1108, 2016.

\bibitem[Pavlakos et~al.(2019)Pavlakos, Choutas, Ghorbani, Bolkart, Osman,
  Tzionas, and Black]{SMPL-X:2019}
Georgios Pavlakos, Vasileios Choutas, Nima Ghorbani, Timo Bolkart, Ahmed A.~A.
  Osman, Dimitrios Tzionas, and Michael~J. Black.
\newblock Expressive body capture: 3d hands, face, and body from a single
  image.
\newblock In \emph{CVPR}, 2019.

\bibitem[Ponomareva et~al.(2021)Ponomareva, Trinitatova, Fedoseev, Kalinov, and
  Tsetserukou]{ponomareva2021grasplook}
Polina Ponomareva, Daria Trinitatova, Aleksey Fedoseev, Ivan Kalinov, and
  Dzmitry Tsetserukou.
\newblock Grasplook: a vr-based telemanipulation system with r-cnn-driven
  augmentation of virtual environment.
\newblock In \emph{2021 20th International Conference on Advanced Robotics
  (ICAR)}, pages 166--171. IEEE, 2021.

\bibitem[Qin et~al.(2021)Qin, Wu, Liu, Jiang, Yang, Fu, and Wang]{qin2021dexmv}
Yuzhe Qin, Yueh-Hua Wu, Shaowei Liu, Hanwen Jiang, Ruihan Yang, Yang Fu, and
  Xiaolong Wang.
\newblock Dexmv: Imitation learning for dexterous manipulation from human
  videos.
\newblock \emph{arXiv preprint arXiv:2108.05877}, 2021.

\bibitem[Qin et~al.(2022)Qin, Su, and Wang]{qin:ral2022}
Yuzhe Qin, Hao Su, and Xiaolong Wang.
\newblock From one hand to multiple hands: Imitation learning for dexterous
  manipulation from single-camera teleoperation.
\newblock \emph{RA-L}, 7\penalty0 (4):\penalty0 10873--10881, 2022.

\bibitem[Qin et~al.(2023)Qin, Huang, Yin, Su, and Wang]{qin2023dexpoint}
Yuzhe Qin, Binghao Huang, Zhao-Heng Yin, Hao Su, and Xiaolong Wang.
\newblock Dexpoint: Generalizable point cloud reinforcement learning for
  sim-to-real dexterous manipulation.
\newblock In \emph{Conference on Robot Learning}, pages 594--605. PMLR, 2023.

\bibitem[Radosavovic et~al.(2022)Radosavovic, Xiao, James, Abbeel, Malik, and
  Darrell]{mvpreal}
Ilija Radosavovic, Tete Xiao, Stephen James, Pieter Abbeel, Jitendra Malik, and
  Trevor Darrell.
\newblock Real-world robot learning with masked visual pre-training.
\newblock \emph{arXiv preprint arXiv:2210.03109}, 2022.

\bibitem[Rajeswaran et~al.(2018)Rajeswaran, Kumar, Gupta, Vezzani, Schulman,
  Todorov, and Levine]{Rajeswaran2018}
Aravind Rajeswaran, Vikash Kumar, Abhishek Gupta, Giulia Vezzani, John
  Schulman, Emanuel Todorov, and Sergey Levine.
\newblock Learning complex dexterous manipulation with deep reinforcement
  learning and demonstrations.
\newblock In \emph{RSS}, 2018.

\bibitem[Rakita et~al.(2017)Rakita, Mutlu, and Gleicher]{rakita2017motion}
Daniel Rakita, Bilge Mutlu, and Michael Gleicher.
\newblock A motion retargeting method for effective mimicry-based teleoperation
  of robot arms.
\newblock In \emph{Proceedings of the 2017 ACM/IEEE International Conference on
  Human-Robot Interaction}, pages 361--370, 2017.

\bibitem[Rakita et~al.(2019)Rakita, Mutlu, and Gleicher]{rakita2019remote}
Daniel Rakita, Bilge Mutlu, and Michael Gleicher.
\newblock Remote telemanipulation with adapting viewpoints in visually complex
  environments.
\newblock \emph{Robotics: Science and Systems XV}, 2019.

\bibitem[Ratliff et~al.(2018)Ratliff, Issac, Kappler, Birchfield, and Fox]{rmp}
Nathan~D Ratliff, Jan Issac, Daniel Kappler, Stan Birchfield, and Dieter Fox.
\newblock Riemannian motion policies.
\newblock \emph{arXiv preprint arXiv:1801.02854}, 2018.

\bibitem[Rong et~al.(2020)Rong, Shiratori, and Joo]{rong2020frankmocap}
Yu~Rong, Takaaki Shiratori, and Hanbyul Joo.
\newblock Frankmocap: Fast monocular 3d hand and body motion capture by
  regression and integration.
\newblock \emph{arXiv preprint arXiv:2008.08324}, 2020.

\bibitem[Rosen et~al.(2020)Rosen, Whitney, Fishman, Ullman, and
  Tellex]{rosen2020mixed}
Eric Rosen, David Whitney, Michael Fishman, Daniel Ullman, and Stefanie Tellex.
\newblock Mixed reality as a bidirectional communication interface for
  human-robot interaction.
\newblock In \emph{2020 IEEE/RSJ International Conference on Intelligent Robots
  and Systems (IROS)}, pages 11431--11438. IEEE, 2020.

\bibitem[Salvato et~al.(2022)Salvato, Heravi, Okamura, and
  Bohg]{salvato2022predicting}
M~Salvato, Negin Heravi, Allison~M Okamura, and Jeannette Bohg.
\newblock Predicting hand-object interaction for improved haptic feedback in
  mixed reality.
\newblock \emph{IEEE Robotics and Automation Letters}, 7\penalty0 (2):\penalty0
  3851--3857, 2022.

\bibitem[Shaw et~al.(2022)Shaw, Bahl, and Pathak]{shaw2022videodex}
Kenneth Shaw, Shikhar Bahl, and Deepak Pathak.
\newblock Videodex: Learning dexterity from internet videos.
\newblock \emph{arXiv preprint arXiv:2212.04498}, 2022.

\bibitem[Sivakumar et~al.(2022)Sivakumar, Shaw, and
  Pathak]{sivakumar2022robotic}
Aravind Sivakumar, Kenneth Shaw, and Deepak Pathak.
\newblock Robotic telekinesis: learning a robotic hand imitator by watching
  humans on youtube.
\newblock \emph{arXiv preprint arXiv:2202.10448}, 2022.

\bibitem[Son et~al.(2013)Son, Franchi, Chuang, Kim, Bulthoff, and
  Giordano]{son2013human}
Hyoung~Il Son, Antonio Franchi, Lewis~L Chuang, Junsuk Kim, Heinrich~H
  Bulthoff, and Paolo~Robuffo Giordano.
\newblock Human-centered design and evaluation of haptic cueing for
  teleoperation of multiple mobile robots.
\newblock \emph{IEEE transactions on cybernetics}, 43\penalty0 (2):\penalty0
  597--609, 2013.

\bibitem[Sundaralingam et~al.(2023)Sundaralingam, Hari, Fishman, Garrett, Wyk,
  Blukis, Millane, Oleynikova, Handa, Ramos, Ratliff, and
  Fox]{Sundaralingam2023CuRobo}
Balakumar Sundaralingam, Siva Hari, Adam Fishman, Caelan Garrett, Karl~Van Wyk,
  Valts Blukis, Alexander Millane, Helen Oleynikova, Ankur Handa, Fabio Ramos,
  Nathan Ratliff, and Dieter Fox.
\newblock {CuRobo}: Parallelized collision-free robot motion generation.
\newblock In \emph{Proceedings of the IEEE International Conference on Robotics
  and Automation ({ICRA})}, June 2023.

\bibitem[Theobalt et~al.(2004)Theobalt, Albrecht, Haber, Magnor, and
  Seidel]{theobalt2004pitching}
Christian Theobalt, Irene Albrecht, J{\"o}rg Haber, Marcus Magnor, and
  Hans-Peter Seidel.
\newblock Pitching a baseball: tracking high-speed motion with multi-exposure
  images.
\newblock In \emph{ACM SIGGRAPH 2004 Papers}, pages 540--547. 2004.

\bibitem[Todorov et~al.(2012)Todorov, Erez, and Tassa]{todorov2012mujoco}
Emanuel Todorov, Tom Erez, and Yuval Tassa.
\newblock Mujoco: A physics engine for model-based control.
\newblock In \emph{IROS}, 2012.

\bibitem[Tung et~al.(2021)Tung, Wong, Mandlekar, Mart{\'\i}n-Mart{\'\i}n, Zhu,
  Fei-Fei, and Savarese]{tung2021learning}
Albert Tung, Josiah Wong, Ajay Mandlekar, Roberto Mart{\'\i}n-Mart{\'\i}n, Yuke
  Zhu, Li~Fei-Fei, and Silvio Savarese.
\newblock Learning multi-arm manipulation through collaborative teleoperation.
\newblock In \emph{2021 IEEE International Conference on Robotics and
  Automation (ICRA)}, pages 9212--9219. IEEE, 2021.

\bibitem[Wang and Popovi{\'c}(2009)]{wang2009real}
Robert~Y Wang and Jovan Popovi{\'c}.
\newblock Real-time hand-tracking with a color glove.
\newblock \emph{ACM transactions on graphics (TOG)}, 28\penalty0 (3):\penalty0
  1--8, 2009.

\bibitem[Wei et~al.(2021)Wei, Huang, and Li]{wei2021multi}
Dong Wei, Bidan Huang, and Qiang Li.
\newblock Multi-view merging for robot teleoperation with virtual reality.
\newblock \emph{IEEE Robotics and Automation Letters}, 6\penalty0 (4):\penalty0
  8537--8544, 2021.

\bibitem[Xiang et~al.(2020)Xiang, Qin, Mo, Xia, Zhu, Liu, Liu, Jiang, Yuan,
  Wang, et~al.]{xiang2020sapien}
Fanbo Xiang, Yuzhe Qin, Kaichun Mo, Yikuan Xia, Hao Zhu, Fangchen Liu, Minghua
  Liu, Hanxiao Jiang, Yifu Yuan, He~Wang, et~al.
\newblock Sapien: A simulated part-based interactive environment.
\newblock In \emph{CVPR}, 2020.

\bibitem[Ye et~al.(2023)Ye, Wang, Huang, Qin, and Wang]{ye2023learning}
Jianglong Ye, Jiashun Wang, Binghao Huang, Yuzhe Qin, and Xiaolong Wang.
\newblock Learning continuous grasping function with a dexterous hand from
  human demonstrations.
\newblock \emph{IEEE Robotics and Automation Letters}, 8\penalty0 (5):\penalty0
  2882--2889, 2023.

\bibitem[Zhang et~al.(2020)Zhang, Bazarevsky, Vakunov, Tkachenka, Sung, Chang,
  and Grundmann]{zhang2020mediapipe}
Fan Zhang, Valentin Bazarevsky, Andrey Vakunov, Andrei Tkachenka, George Sung,
  Chuo-Ling Chang, and Matthias Grundmann.
\newblock Mediapipe hands: On-device real-time hand tracking.
\newblock \emph{arXiv preprint arXiv:2006.10214}, 2020.

\bibitem[Zhang et~al.(2021)Zhang, Li, Liang, Chen, Cui, Wang, and
  Xiong]{zhang2021human}
Haodong Zhang, Weijie Li, Yuwei Liang, Zexi Chen, Yuxiang Cui, Yue Wang, and
  Rong Xiong.
\newblock Human-robot motion retargeting via neural latent optimization.
\newblock \emph{CoRR}, 2021.

\bibitem[Zhang et~al.(2018{\natexlab{a}})Zhang, Zhao, Yu, Gui, Sheng, and
  Zhu]{zhang2018feasibility}
Heng Zhang, Zeming Zhao, Yang Yu, Kai Gui, Xinjun Sheng, and Xiangyang Zhu.
\newblock A feasibility study on an intuitive teleoperation system combining
  imu with semg sensors.
\newblock In \emph{Intelligent Robotics and Applications: 11th International
  Conference, ICIRA 2018, Newcastle, NSW, Australia, August 9--11, 2018,
  Proceedings, Part I 11}, pages 465--474. Springer, 2018{\natexlab{a}}.

\bibitem[Zhang et~al.(2018{\natexlab{b}})Zhang, McCarthy, Jow, Lee, Chen,
  Goldberg, and Abbeel]{zhang2018deep}
Tianhao Zhang, Zoe McCarthy, Owen Jow, Dennis Lee, Xi~Chen, Ken Goldberg, and
  Pieter Abbeel.
\newblock Deep imitation learning for complex manipulation tasks from virtual
  reality teleoperation.
\newblock In \emph{ICRA}, 2018{\natexlab{b}}.

\bibitem[Zhao et~al.(2012)Zhao, Chai, and Xu]{zhao2012combining}
Wenping Zhao, Jinxiang Chai, and Ying-Qing Xu.
\newblock Combining marker-based mocap and rgb-d camera for acquiring
  high-fidelity hand motion data.
\newblock In \emph{Proceedings of the ACM SIGGRAPH/eurographics symposium on
  computer animation}, pages 33--42, 2012.

\end{thebibliography}
